\documentclass[lettersize,journal]{IEEEtran}
\usepackage{amsmath,amsfonts}
\usepackage{array}
\usepackage{textcomp}
\usepackage{stfloats}
\usepackage{url}
\usepackage{verbatim}
\usepackage{graphicx}
\usepackage{cite}
\usepackage{booktabs}
\usepackage{color}
\usepackage{bbm}
\usepackage{bm}
\usepackage{amsmath}
\usepackage{multirow}
\usepackage{algorithm}
\usepackage{algpseudocode}
\usepackage{subfigure}
\usepackage{wrapfig}
\usepackage[table]{xcolor}
\usepackage[framemethod=tikz]{mdframed}
\usepackage{lipsum}

\newcommand{\param}{{\bm{\theta}}}
\newcommand{\setB}{{\mathcal{B}}}

\definecolor{Teal}{RGB}{51, 142, 167}
\definecolor{Red}{RGB}{0, 0, 0}
\definecolor{Blue}{RGB}{0, 0, 0}
\usepackage[colorlinks,
            linkcolor=blue,
            anchorcolor=blue,
            citecolor=blue,
]{hyperref}

\begin{document}

\title{
  Enhancing Sample Utilization in\\
  Noise-robust Deep Metric Learning with\\
  Subgroup-based Positive-pair Selection
}

\author{\IEEEauthorblockN{Zhipeng Yu,
    Qianqian Xu, \emph{Senior Member, IEEE,}
    Yangbangyan Jiang\IEEEauthorrefmark{2}, \emph{Member, IEEE,}\\
    Yingfei Sun\IEEEauthorrefmark{2},
    and Qingming Huang}, \emph{Fellow, IEEE}

  \par\bigskip
  \thanks{
    Z. Yu and Y. Sun are with the School of Electronic, Electrical and Communication Engineering, University of Chinese Academy of Sciences,
    Beijing 100049, China (E-mail: yuzhipeng21@mails.ucas.ac.cn, yfsun@ucas.ac.cn).

    Q. Xu is with the Key Laboratory of Intelligent Information Processing, Institute of Computing Technology, Chinese Academy of Sciences, Beijing 100190, China (E-mail: xuqianqian@ict.ac.cn).

    Y. Jiang is with the School of Computer Science and Technology, University of Chinese Academy of Sciences, Beijing 101408, China (E-mail: jiangyangbangyan@ucas.ac.cn).

    Q. Huang is with the School of Computer Science and Technology, University of Chinese Academy of Sciences, Beijing 101408, China, also with the Key Laboratory of Intelligent Information Processing, Institute of Computing Technology, Chinese Academy of Sciences, Beijing
    100190, China (E-mail: qmhuang@ucas.ac.cn).

    \IEEEauthorrefmark{2} Corresponding authors.
  }
}
\markboth{IEEE TRANSACTIONS ON IMAGE PROCESSING, VOL. 33, 2024}%
{Shell \MakeLowercase{\textit{et al.}}: A Sample Article Using IEEEtran.cls for IEEE Journals}

\maketitle
\begin{abstract}
  The existence of noisy labels in real-world data negatively impacts the performance of deep learning models. Although much research effort has been devoted to improving the robustness towards noisy labels in classification tasks, the problem of noisy labels in deep metric learning (DML) remains under-explored. Existing noisy label learning methods designed for DML mainly discard suspicious noisy samples, resulting in a waste of the training data. To address this issue, we propose a noise-robust DML framework with SubGroup-based Positive-pair Selection (SGPS), which constructs reliable positive pairs for noisy samples to enhance the sample utilization. Specifically, SGPS first effectively identifies clean and noisy samples by a probability-based clean sample selectionstrategy. To further utilize the remaining noisy samples, we discover their potential similar samples based on the subgroup information given by a subgroup generation module and then aggregate them into informative positive prototypes for each noisy sample via a positive prototype generation module. Afterward, a new contrastive loss is tailored for the noisy samples with their selected positive pairs. SGPS can be easily integrated into the training process of existing pair-wise DML tasks, like image retrieval and face recognition. Extensive experiments on multiple synthetic and real-world large-scale label noise datasets demonstrate the effectiveness of our proposed method. Without any bells and whistles, our SGPS framework outperforms the state-of-the-art noisy label DML methods. Code is available at \url{https://github.com/smuelpeng/SGPS-NoiseFreeDML}.
\end{abstract}


\section{Introduction}
\IEEEPARstart{D}{eep} learning has achieved remarkable success in various computer vision domains,
including image classification~\cite{he2016deep}, image retrieval (IR)~\cite{wang2019multi}, and face recognition (FR)~\cite{deng2019arcface}.
Due to the powerful representation ability,
deep neural networks can extract useful information from large-scale annotated datasets.
However, human annotations or auto-collected data will inevitably introduce
unexpected noise in real-world datasets.
The ubiquitous noisy labels might largely degrade the performance of deep learning models.

\begin{figure*}
  \centering
  \subfigure[]{
    \includegraphics[width=0.31\linewidth]{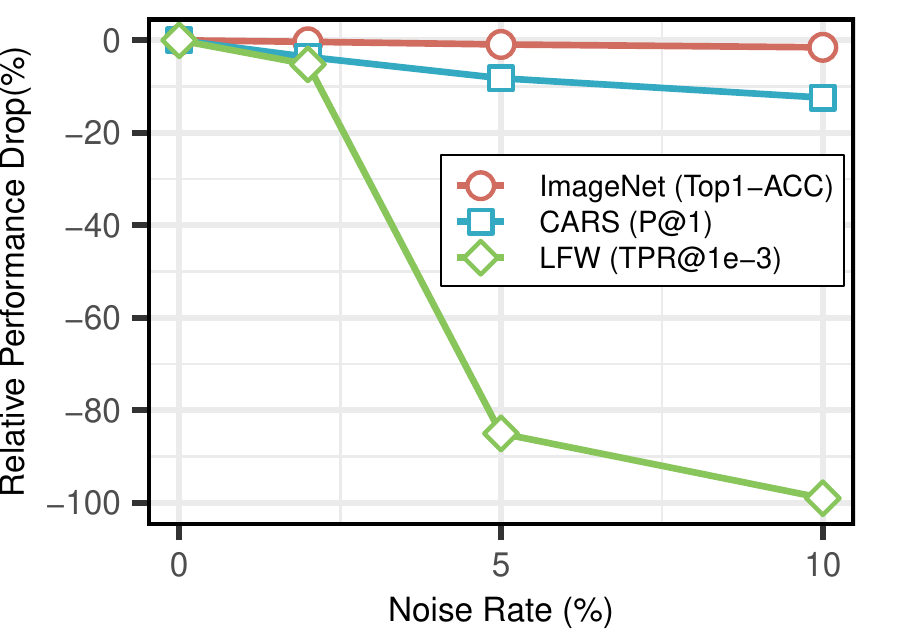}
    \label{fig:intro_a}
  }
  \subfigure[]{
    \includegraphics[width=0.31\linewidth]{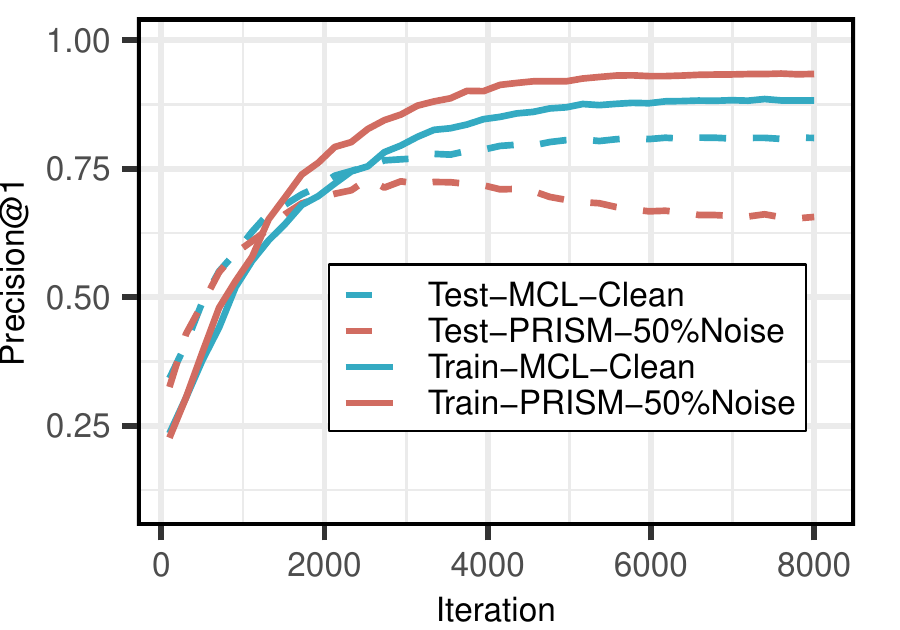}
    \label{fig:intro_b}
  }
  \subfigure[]{
    \includegraphics[width=0.31\linewidth]{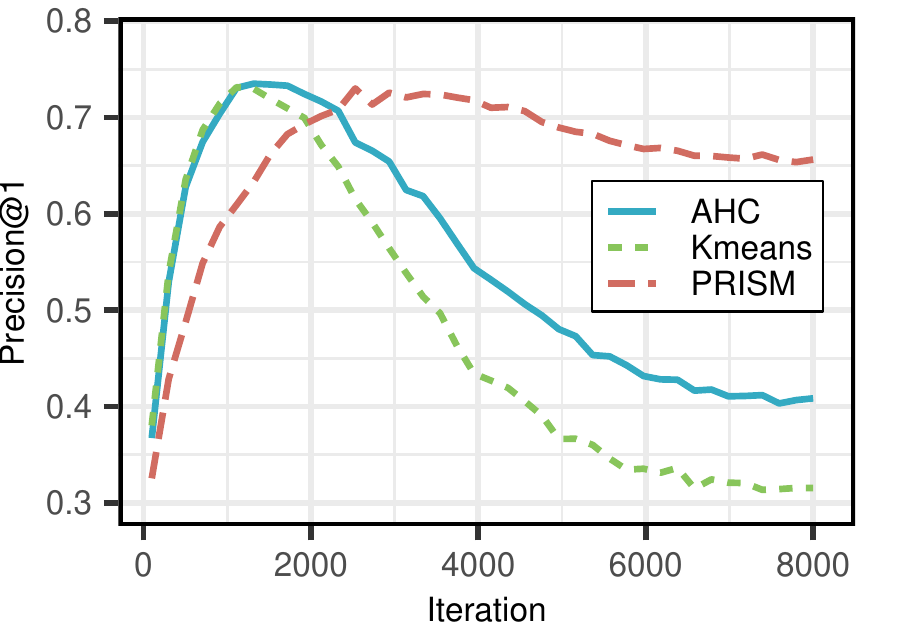}
    \label{fig:intro_c}
  }
  \vspace{-2mm}
  \caption{
    (a) Performance drop corresponding to the noise level for
    image classification and DML tasks
    (image retrieval and face recognition).
    (b) Precision@1 of the clean training subset and test set for
    different methods on CARS.
    (c) Precision@1 of PRISIM~\cite{PRISM2021} and
    clustering-based denoising methods, i.e., Kmeans~\cite{kmeans++} and
    agglomerative hierarchical clustering (AHC)~\cite{nielsen2016hierarchical}
    on CARS with 50\% sysmmetric noise.
  }
  \vspace{-5mm}
  \label{fig:intro}
\end{figure*}

Recent noisy label learning research has made great progress on
classification tasks through various methods,
such as sample selection~\cite{han2018co,xia2022sample,li2022selective, zhao2022dist},
weight generation~\cite{jiang2018mentornet,zheng2021meta},
transfer matrix~\cite{F-correction,Li2022MLT, yang2023CLTM},
and semi-supervised methods~\cite{li2020dividemix,bai2021memomentum,jiang2022maxmatch}.
Nevertheless, little effort has been devoted to the problem of noisy labels in deep metric learning (DML).
In DML, the test set usually consists of classes that are not present in the training set.
Moreover, DML sorely relies on the similarity between sample features to determine sample relationships
instead of the probability prediction of classifiers.
Therefore, many DML tasks prefer to use pair-wise loss functions
~\cite{chopra2005learning, wang2019multi,wang2020cross} to optimize the model,
which makes it difficult to directly apply the classifier-based noisy label learning methods. What's worse, DML model is more sensitive to label noise than basic classification ones~\cite{dereka2022deep}. As shown in Fig.\ref{fig:intro_a}, we can observe a rapid performance drop with the increase of uniform label noise level on retrieval tasks. Specifically, even 5\% of label noise could cause a relative performance drop of nearly 10\% on image retrieval (CARS) and more than 80\% on face recognition (LFW) tasks. In contrast, the classification accuracy on ImageNet under the same
condition only exhibits a much smaller drop (less than 2\%).

The current state-of-the-art noise label learning method for DML is PRISM~\cite{PRISM2021}, which mainly relies on prototype similarity ranking to select clean samples and construct reliable sample pairs for DML training. However, PRISM tends to discard excessive noisy samples, resulting in a prevalent overfitting issue.
As shown in Fig.\ref{fig:intro_b}, even in the presence of noisy labels, PRISM could outperform the baseline MCL model trained on the original clean data in terms of the Precision@1 on the clean training subset used in PRISM. However, on the test set, we can observe a substantial performance drop for PRISM during later training stages. This phenomenon can be attributed to confirmation bias, where the model accumulates confidence for the selected clean samples and fails to explore potentially valuable noisy samples continually. Therefore, it is urgent to utilize the ignored noisy samples to benefit the noisy label DML.

The most direct way to involve the noisy samples is to learn with the pseudo-labels generated by unsupervised methods like clustering. Nevertheless, clustering-based methods are not immune to the overfitting problem. We apply K-means~\cite{kmeans++} and agglomerative hierarchical clustering (AHC)~\cite{nielsen2016hierarchical} to generate pseudo-labels from the features extracted by an early-stopped PRISM model. Then new models are trained using these pseudo-labels from scratch. From Fig.\ref{fig:intro_c}, although the performance is improved at the very beginning, it eventually suffers from significant degradation. This decline is primarily attributed to the additional error introduced in cluster labels, whose accuracy is significantly constrained by the quality of the pretrained features.

In fact, from the perspective of pairwise learning, the noisy labels (also including pseudo-labels) will incur both wrong positive and negative pairs. In DML tasks, positive pairs often play an more important role than negative ones. Motivated by this, we turn our attention to construct more informative sample pairs, especially the positive ones, for noisy samples. We propose a noise-robust DML training framework which enhances the sample utilization by \textbf{S}ub\textbf{G}roup-based \textbf{P}ositive pair \textbf{S}election (\textbf{SGPS}). The whole framework is illustrated in Fig.\ref{fig:method_frame}.
Without need the predictions of classifiers, SGPS introduces a probability-based clean sample selection (PCS) strategy to effectively identify the clean and noisy samples. We simply use the historical features stored in the memory bank to calculate the probability of each sample being clean. The memory bank works as a first-in-first-out queue with a limited size which only store latest sample features.
Besides learning with clean samples using general DML objectives (denoted as $L_\text{clean}$), SGPS goes a step further by discovering extra positive pairs for noisy samples based on the subgroup labels obtained by a subgroup generation module (SGM). Another feature bank is used for subgroup generation in SGM, which stores the features of all training samples and is updated in a momentum way. Compared with cluster labels, our subgroup labels integrate the knowledge from both the original labels and the feature distribution, reducing the incorrect positive and negative pairs. After that, a positive prototype generation module (PPM) is applied to aggregate multiple positive samples for each noisy sample into positive prototypes. And then these aggregated pairs is used to compute a noisy contrastive loss function $L_\text{noise}$ for noisy samples. By adding $L_\text{clean}$ and $L_\text{noise}$ together, both clean and noisy samples can be well utilized for DML training.

The contributions of this paper are summarized as follows:
\begin{itemize}
  \item We propose a new noise-robust training framework for DML with SubGroup-based Positive pair Selection (SGPS). Unlike existing methods, SGPS integrates both clean and noisy samples into the training process to enhance the sample utilization.
  \item In the core of the framework, we introduce SGM to efficiently generate subgroup labels for noisy samples. The subgroup information helps the discovery of true positive pairs without imposing significant time costs during training.
  \item We then present PPM to aggregate multiple positive samples into a prototype for each noisy samples, leading to more informative positive pairs.
\end{itemize}
Finally, we provide extensive experimental results to demonstrate the effectiveness of our proposed method. SGPS can be plugged into both existing proxy- and pair-based DML modles to improve performance.

\section{Related work}
\subsection{Metric Learning}
We broadly categorize deep metric learning into 1) pair-wise and 2) proxy-based methods. Pair-wise methods \cite{schroff2015facenet},~\cite{wang2019multi,sun2020circle,zhang2022attributable} calculate the loss based on the
contrast between positive pairs and negative pairs. Commonly used loss functions include contrastive loss \cite{chopra2005learning}, triplet loss~\cite{schroff2015facenet} and softmax loss~\cite{goldberger2004neighbourhood}. In this process, identifying informative positive and negative pairs becomes an important consideration,~\cite{wang2019multi}. The problem of instance sampling in DML has also been considered~\cite{wu2017sampling}.
{Nevertheless, objectives like triplet loss is in fact optimising a distance metric rather than a ranking metric, which is sub-optimal when evaluating using a ranking metric, such as Average Precision (AP) and Area Under the Precision-Recall Curve (AUPRC).  Therefore, a series of ranking-based methods have been proposed to address the shortcomings of distance-based pair-wise methods, enabling neural networks to achieve end-to-end training to directly optimize the ranking metrics.
As a typical approach, FastAP~\cite{FastAp} optimizes the AP metric using an approximation derived from distance quantization.
Roadmap~\cite{ramzi2021robust} proposes robust and decomposable objective to address non-differentiability and non-decomposability challenges for end-to-end training of deep neural networks with AP.
\cite{wen2022auroc} proposes a sampling-rate-invariant unbiased stochastic estimator with superior stability to optimize AUPRC.}
On the other hand, proxy-based methods~\cite{nSoftmax, proxyNCA, qian2019softtriple},~\cite{teh2020proxynca++, an2023unicom} represent each class with one or more proxy vectors, and use the similarities between the input data and the proxies to calculate the loss. Proxies are learned from data during model training, which could deviate from the class center under heavy noise and cause performance degradation. {Recently, Contextual \cite{liao2023supervised} introduces a contextual loss to learn the feature representation and the similarity metric jointly.
    Some methods like ~\cite{sun2020circle, yan2022metricformer} try to integrate pair-wise and proxy-based approaches into a unified framework. MetricFormer \cite{yan2022metricformer} considers the correlations from a unified perspective to capture and models the multiple correlations with an elaborate metric transformer.
    Additionally, CIIM \cite{yan2024causality} proposes a causality-invariant interactive mining method to learn the cross-modal similarity with the causal structure by modality-aware hard mining and modality-invariant feature embedding for cross-modal image DML tasks.}

\subsection{Noisy Label Learning}
Training under noisy labels has been studied extensively for classification tasks~\cite{angluin1988learning, wang2018iterative, zheng2021meta,yao2023better,ni2023psnea, zhao2022dist, jiang2023positive}. A common approach is to detect noisy labels and exclude them from the training set. Recent works have also started exploring correcting the noisy labels~\cite{angluin1988learning,zheng2021meta}or treating noisy data as unlabeled data for semi-supervised learning~\cite{li2020dividemix}. F-correction~\cite{F-correction} models label noise with a class transition matrix. MentorNet~\cite{jiang2018mentornet} trains a teacher network that provides a weight for each sample to the student network. Co-teaching~\cite{han2018co} trains two networks concurrently. Samples identified as having small losses by one network are used to train the other network. This is further improved in~\cite{coteaching+} by training on samples with small losses and different predictions from the two networks.
Many recent works~\cite{li2020dividemix,nishi2021improving,jiang2022maxmatch,wang2024regularized,jiang2023positive} filter noisy data based on the predicted class output by the current classification model and introduce semi-supervised learning to improve the robustness of the model. {Among them, Unicon \cite{karim2022unicon} employs a semi-supervised approach with class balance constraints, enhancing the performance of classical semi-supervised based methods like DivideMix \cite{li2020dividemix}.
    Sel-CL \cite{li2022selective} is based on selective-supervised contrastive learning and reliable sample and pairs selection.
    Seeing the recent advances in the diffusion model, LRA-Diffusion \cite{chen2024label} attempts to combine pretrained features and the diffusion model to infer real labels directly and achieves the state-of-the-art performance.
    Early stopping \cite{bai2021pes, yuanearly} acts as a simple yet effective method to reduce the impact of noisy samples by stopping the training process when the model starts to overfit the noisy samples.
    Specifically, LabelWave \cite{yuanearly} simplifies the early stopping method which does not require validation data for selecting the desired model in the presence of label noise.
    These studies could exhibit a good label-noise robustness on the image classification datasets.
  }
Besides, unsupervised clustering-based methods ~\cite{kmeans++,nielsen2016hierarchical, jiang2019dm2c,jiang2018learn,jiang2019duet} are also widely used to generate pseudo-labels for noisy samples.

\subsection{Noisy Labels in Metric Learning}
Previous work on noisy labels in metric learning is limited.~\cite{wang2017robust} estimates the posterior label distribution
using a shallow Bayesian neural network, which may not scale well to deeper network architectures.~\cite{wang2019deep} uses the pair-wise loss and trains a proxy for each class simultaneously to adjust the weights of the outliers. In~\cite{ozaki2019large}, noisy data in DML is handled by first performing label cleaning using a model trained on a clean dataset, which may not be available in real-world applications. Recently, methods based on hyperbolic embedding space~\cite{ermolov2022hyperbolic,yan2023adaptive} and hierarchical labels~\cite{yan2021unsupervised} have been proposed to improve the robustness of metric learning on unsupervised and noisy label settings.
  {There are also many other ways to reduce the impact of noisy samples.
    For example, One4More\cite{zhang2021one} learns a data sampler to reduce the sampling frequency on noisy samples.
    RTMem \cite{ye2021collaborative} updates the cluster centroid with a randomly sampled instance feature in the current mini-batch without momentum to reduce the mismatching between the model and feature memory.
    LaCoL \cite{yan2022noise} proposes latent contrastive learning to leverage the negative correlations from the noisy data to guarantee the robust generalization of DML models.
    LP \cite{lan2023learning} introduces multi-view features and teacher-student distillation to purify noise in pseudo labels.
    PRISM~\cite{PRISM2021} uses memory features of the same category to identify noisy samples and select clean samples to construct reliable sample pairs for DML training, which achieves state-of-the-art performance on noisy DML tasks.
    Our work shares a similar motivation to the PRISM method.
    Nevertheless, we resample positive pairs for noisy samples instead of discarding them.}

\section{Approach}
\begin{figure*}[!htp]
  \begin{center}
    \includegraphics[width=0.93\linewidth]{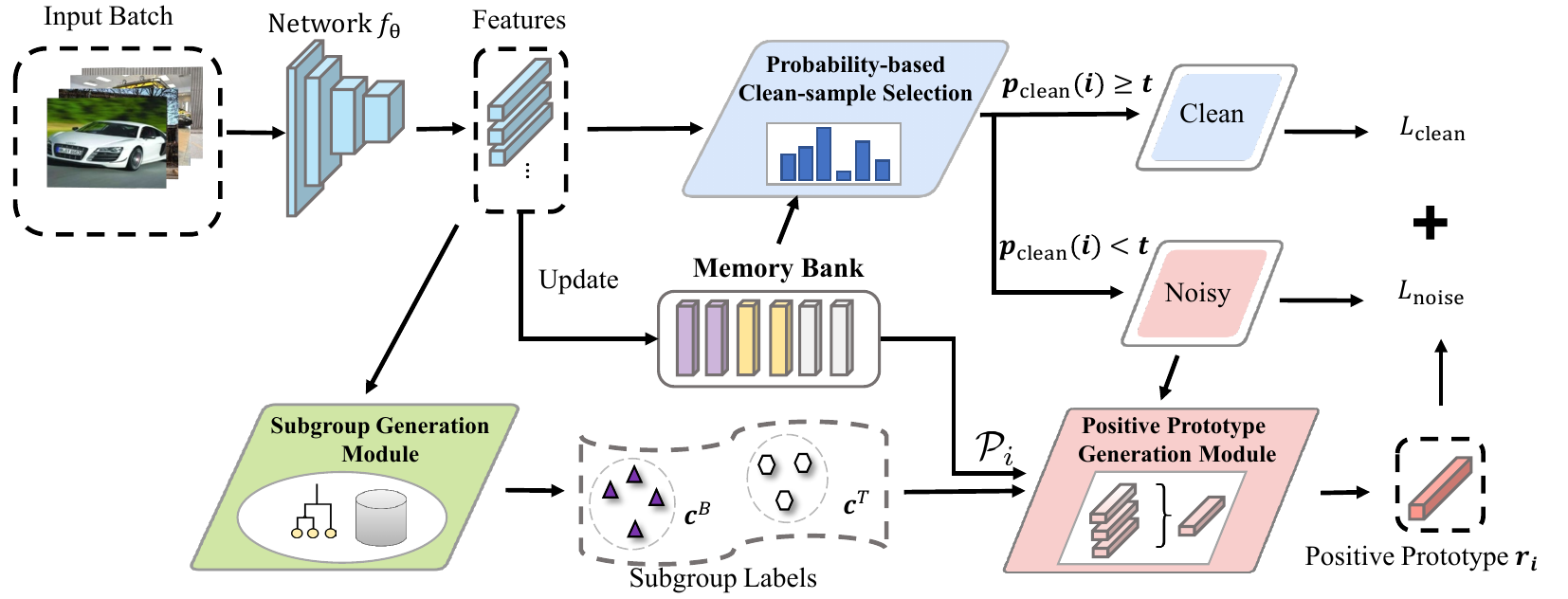}
    \caption{The framework of our proposed method. The input batch will first be fed into the feature extractor network to obtain the features. Then, inputs will be separated into a clean set and a noisy set by the PCS module. Samples in the clean set will be used to compute $L_{\text{clean}}$. Based on the subroup labels generated by SGM, samples in the noisy set also obtain corresponding positive pairs $\mathcal{P}_{i}$. PPM will aggregate $\mathcal{P}_{i}$ to generate positive prototype $\bm{r}_{i}$ to compute $L_{\text{noise}}$ with the noisy samples. }
    \label{fig:method_frame}
  \end{center}
\end{figure*}

\subsection{Problem Setting}
Let $\{(\bm{x}_i, y_i)\}^N_{i=1}$ denote the set of $N$ training samples and the corresponding annotated labels from $M$ classes. The $d$-dimensional features are extracted by a neural network $f$ parameterized by $\param$. Define the similarity between two samples as $S(f_\param(\bm{x}_i), f_\param(\bm{x}_j)) = \frac{f_\param(\bm{x}_i)^\top f_\param(\bm{x}_j)}{\|f_\param(\bm{x}_i)\| \cdot \|f_\param(\bm{x}_j)\|}$ (denoted as $S_{i j}$ for short). Then, the objective of DML is to maximize the similarity between positive pairs (sample pairs with the same label) and minimize the similarity between negative pairs (sample pairs with different labels). Note that in DML, the positive pairs are usually much less than the negative ones and play a more critical role in learning. Recent pair-wise DML methods such as MS loss~\cite{wang2019multi} and Circle loss~\cite{sun2020circle} well capture pairwise similarity information by adjusting different weights or assigning different margins for positive and negative pairs. However, existing DML methods are sensitive to label noise. The incurred false positive and negative pairs, especially the former, could lead to a sharp performance drop.

Current methods for handling label noise in DML, such as PRISM~\cite{PRISM2021}, mainly focus on filtering clean samples for training. For each sample $\bm{x}_i$, they calculate a confidence score $p_{\text{clean}}(i)$ of how likely the sample is correctly labeled. If $p_{\text{clean}}(i) > t$ where $t$ is a threshold that changes during training, the sample is added into the clean-labeled subset $\setB_{\text{clean}}$. Then DML losses such as the contrastive loss~\cite{chopra2005learning} can be applied to $\setB_{\text{clean}}$. A typical shortcoming of this type of method is that they might discard a substantial number of samples, potentially fostering confirmation bias. Therefore, we propose to discover more positive pairs from the filtered-out noisy samples for learning.

\subsection{Overview}
An overview of the proposed SGPS framework is illustrated in Fig.~\ref{fig:method_frame}. For each batch, we first perform probability-based sample selection to divide samples in a batch into a clean-labeled subset $\setB_{\text{clean}}$ and a noisy-labeled subset $\setB_{\text{noise}}$.
For $\setB_{\text{clean}}$, we can directly compute a common-used DML loss $L_{\text{clean}}$.
Both Pair-wise and Proxy-based DML losses can be selected as $L_{\text{clean}}$.
Moreover, for each instance in $\setB_{\text{noise}}$, we obtain additional positive samples lying from the corresponding subgroup using the SGM.
In this way, we can obtain a set of positive pairs regardless of the original annotated labels in $\setB_{\text{noise}}$. To better utilize the selected positive pairs, a PPM is designed to aggregate the positive pairs to generate positive prototypes for noisy samples. Finally, we can compute the $L_{\text{noise}}$ based on $\setB_{\text{noise}}$.

\subsection{Probability-based Clean-sample Selection}
Inspired by PRISM~\cite{PRISM2021}, we adopt a probability-based clean-sample selection (PCS) strategy to identify clean samples in a mini-batch, together with a memory bank storing historic sample features. Note that the class membership could be indicated by the similarity between a sample and all class centroids. If a sample is the most similar to the class consistent with its annotation, then it is likely to be clean. Therefore, based on the similarity to each class centroid calculated using the memory bank, we could compute the probability that a sample's label is clean as follows:
\begin{equation}
  \begin{aligned}
    p_{\text{clean}}(i)=\exp \left(w_{y_i} \frac{f_\param(\bm{x}_i)}{\left\|f_\param(\bm{x}_i)\right\|}\right) / \sum_{m \in [M]} \exp \left(w_m \frac{f_\param(\bm{x}_i)}{\left\|f_\param(\bm{x}_i)\right\|}\right),
  \end{aligned}
\end{equation}
where $w_{m}$ is the centroid of the $m$-th class obtained from the memory bank. This may be seen as modeling the posterior probability of the data label. A smooth top-R (sTRM) trick ~\cite{PRISM2021} is also adopted to adjust the threshold $t$ for noisy data identification. Let $Q_j$ be the $R$-th percentile of $p_{\text{clean}}(i)$ value in $j$-th mini-batch. Then the threshold $t$ is defined as:
\begin{equation}
  \begin{aligned}
    t=\frac{1}{\omega} \sum_{j=\text{iter}-\omega}^{\text{iter}} Q_j
  \end{aligned}
\end{equation}
where $\omega$ is the sliding window size of training iterations, $R$ is a predefined noise ratio. For data points with a high probability to be clean, i.e., $p_{\text{clean}}(i) \geq  t$, a common DML loss $L_{\text{clean}}$ will be calculated, and their features will also be used to update the memory bank.

\subsection{Subgroup Generation Module}
\begin{figure}[!t]
  \begin{center}
    \includegraphics[width=1.0\linewidth]{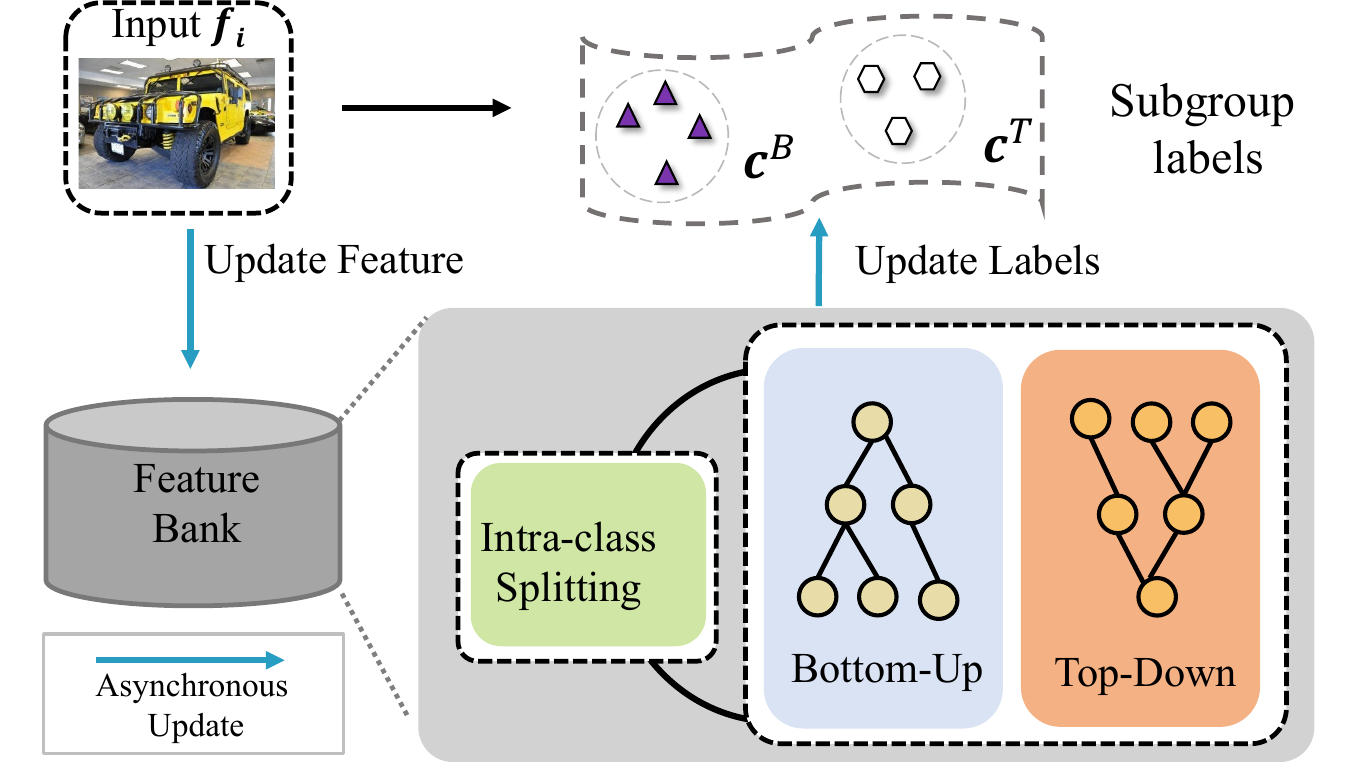}
    \caption{The workflow of SGM.
      SGM will maintain two kinds of subgroup labels for each sample in the dataset.
      The subgrouping module will update the subgroup labels conditionally once new features are added to the feature bank,
      then update the subgroup labels in an asynchronous manner.
      The subgroup labels will be used to select positive pairs for each noisy sample in a batch.
    }
    \label{fig:sgm}
  \end{center}
\end{figure}

The subgroup generation module (SGM) is designed to generate high-confidence candidate positive pairs for $\setB_\text{noise}$. SGM works as a backend server consisting of four parts: (1) feature bank, (2) intra-class splitting submodule, (3) bottom-up subgroup generation submodule, and (4) complementary top-down subgroup generation submodule. The wrongly-annotated labels will incur two types of incorrect pairs in DML:
\begin{itemize}
  \item false negative pairs, in which the samples annotated as different classes actually belong to the same ground-truth category;
  \item false positive pairs, where the samples with the same annotated labels belong to different ground-truth categories.
\end{itemize}
Therefore, we first design a strategy to mitigate the false positive pairs by splitting the noisy samples labeled as the same category into multiple cleaner subgroups. As for false negative pairs, we propose two sub-cluster merging strategies: bottom-up subgroup generation and complementary top-down subgroup generation. The obtained two kinds of subgroup label information will be used to generate positive pairs for each noisy sample. The whole pipeline of SGM is shown in Fig.\ref{fig:sgm}.

\noindent\textbf{Feature Bank.}
Motivated by previous work like MoCo~\cite{Moco} and ODC~\cite{ODC}, we maintain a feature bank that stores features and labels of the entire dataset. Each time a sample is extracted by the network forward pass with L2 normalization, its feature is used to update the sample memory:
\begin{equation}
  \begin{aligned}
    F_{m}(\bm{x}_i) \leftarrow \alpha \frac{f_\param(\bm{x}_i)}{\lVert f_\param(\bm{x}_i)\rVert_2} + (1-\alpha) F_m(\bm{x}_i),
    \label{eq:feature_update}
  \end{aligned}
\end{equation}
where $\alpha$ is a momentum parameter. Considering the large scale of training data in current DML tasks, in order to conserve the training resources and speed up the training process, we deploy the feature bank in additional training nodes distributionally.

\noindent\textbf{Intra-class splitting.}
This submodule aims at mitigating the false positive pairs by splitting samples labeled as the same category into multiple cleaner subgroups. Samples within the same subgroup are more likely to belong to the same real category. Specifically, we first calculate the intra-class similarity matrix $\bm{S}^{m}$ for all samples annotated as the $m$-th class. For the $i$-th sample, we select its 1-nearest-neighbor and other samples very close to it as candidates that might belong to the same subgroup. This subsequently constructs an intra-class adjacency matrix $\bm{W}^m$:
\begin{equation}
  \begin{aligned}
    W^m_{ij} = \begin{cases}
                 1, & \text { if } S^m_{ij} = \max_k S^m_{ik}  \text { or } S^m_{ij} > \lambda_{max} \\
                 0, & \text { otherwise }\end{cases},
    \label{eq:adjacency_matrix}
  \end{aligned}
\end{equation}
where $\lambda_{max}$ is a predefined threshold. Besides, we further set a lower bound $\lambda_{min}$ for $S^m_{ij}$ to eliminate outliers. We then apply a common Connected Components Labeling (CCL)~\cite{CCL} process to obtain subgroups $\{\mathcal{C}_i^{m}\}$. Here, we refer to the subgroup with the largest number of samples for each class $m$ as its meta cluster, denoted as $\mathcal{C}^{m\star}$. The whole pipeline is shown in Algorithm~\ref{alg:intra_cluster}. After this process, we can obtain a set of subgroups for each category $m$. However, directly using these subgroups to build positive pairs is still insufficient. On the other hand, the total number of subgroups is much larger than that of categories $M$, and different subgroups may still belong to the same category, resulting in extra false negative pairs.

\begin{algorithm}[t]
  \caption{Intra Class Splitting}
  \label{alg:intra_cluster}
  \begin{algorithmic}[1]
    \Require
    Training set $\mathcal{S} = \{(\bm{x}_i, y_i)\}_{i}^{N}$;
    The lower and upper similarity thresholds $\lambda_{min}, \lambda_{max}$
    \For{each class $m \in [M]$}
    \State Select all $x_i$ with $y_i = m$ from $\mathcal{S}$
    \State Calculate the corresponding similarity matrix $\bm{S}^{m}$
    \State Construct the adjacency matrix $\bm{W}^{m}$ based on Eq.~\eqref{eq:adjacency_matrix}
    \State Set $W_{ij}^{m} = 0$ if $S_{ij}^{m} < \lambda_{min}$ \Comment{\textit{Remove outliers}}
    \State Acquire the subgroups $\{\mathcal{C}_i^{m}\}$ based on the connected components of $\bm{W}^{m}$
    \State Choose the meta cluster $\mathcal{C}^{m\star} = \arg\max_i | \mathcal{C}_i^{m} |$
    \EndFor
    \Ensure All the subgroups $\mathcal{C} = \bigcup^M_m \mathcal{C}^{m}$
  \end{algorithmic}
\end{algorithm}

\noindent\textbf{Bottom-up subgroup generation.}
Initially, each small subgroup is treated as a separate clean cluster, resulting in a total of $M_c$ clusters. We first calculate the similarity between each pair of cluster centroids:
\begin{equation}
  \begin{aligned}
    \bar{S}_{ij} =   \frac{\bar{\bm{f}}_i ^{\,\top}  \cdot \bar{\bm{f}}_j}{\|\bar{\bm{f}}_i\|  \cdot
    \|\bar{\bm{f}}_j\|},
    \label{eq:cluster_similarity}
  \end{aligned}
\end{equation}
where $\bar{\bm{f}}_i$ is the centroid (mean feature) of cluster $i$. Following the hierarchical bottom-up subgrouping approach~\cite{johnson1967hierarchical,comaniciu2002mean}, we merge the two most similar clusters at each step and
then recalculate the similarity between the merging cluster and other clusters. The number of clusters will decrease by one after each merge step. This merging process continues until the number of clusters is no more than $\tau_{k}$ or the similarity between the two clusters is less than $\lambda'_{min}$. The whole process is shown in Algorithm \ref{alg:BT_cluster}. During this process, potential disruptions might be caused by highly-ranked noise clusters.
Such noisy clusters are often highly similar to other clusters, which may result in a supercluster containing multiple categories. To tackle this problem, we establish a set of rules to avoid over-merging:
\begin{itemize}
  \item Meta group exclusion: A meta cluster is not allowed to be merged with other meta clusters unless their similarity is greater than $\lambda'_{max}$.
  \item Class size control: The number of samples in the merged cluster cannot exceed $\tau_{max}$.
\end{itemize}
At the end of bottom-up subgroup generation, We can obtain the bottom-up subgroup labels $\bm{c}^{B}$.
\begin{algorithm}[t]
  \caption{Bottom-up subgroup generation }
  \label{alg:BT_cluster}
  \begin{algorithmic}[1]
    \Require
    Initial subgroups $\{\mathcal{C}_i\}^{M_c}_{i=1}$;
    The number of minimum subgroups after merging $\tau_{k}$;
    The maximum sample number in each subgroup $\tau_{max}$;
    The thresholds $\lambda'_{min}$;
    \State Set $M_c^{'} = M_{c}$
    \State Compute the centroid $\bar{\bm{f}}_i$ of each subgroup
    \State Compute the similarity between each pair of cluster centroids $\bar{S}_{ij}$
    \Repeat
    \State Find two subgroups $\mathcal{C}_i, \mathcal{C}_j$ with the largest similarity
    \If {not both $\mathcal{C}_i$ and $\mathcal{C}_j$ are meta clusters}
    \If {$\vert \mathcal{C}_i \vert + \vert \mathcal{C}_j \vert < \tau_{max}$}
    \State Merge $\mathcal{C}_i, \mathcal{C}_j$ into a new subgroup $\mathcal{C}_h$
    \State Compute similarity between $\mathcal{C}_h$ and other subgroups
    \State Set $M_{c}^{'} = M_{c}^{'} - 1$
    \EndIf
    \EndIf
    \State Set $\bar{S}_{ij} = 0$ \Comment{Ignore the current subgroup pair}
    \Until{$M_c' < \tau_k \vee \forall i, j: \bar{S}_{ij} < \lambda'_{\text{min}}$}
    \Ensure Subgroup labels for each sample $\bm{c}^{B} = \{c_i^{B}\}_{i=1}^N$
  \end{algorithmic}
\end{algorithm}

\noindent\textbf{Complementary top-down subgroup generation.}
Bottom-up merging procedure heavily relies on the feature quality and the density distribution of samples. Once high-similarity noisy clusters are merged, more noisy samples will be included in the merged cluster, which finally results in a super-large group containing multiple categories. Positive groups with low similarity are often unable to merge effectively,
leading to the omission of many potential positive samples. To address this issue, we propose a complementary top-down subgroup generation procedure to discover more underlying positive samples. In this method, we first assume that all clusters belong to one cell, then apply a recursive partitioning method to divide the entire cluster vector space into small cells.
In each iteration, we select a cell containing more than $B$ samples, then randomly select a pair of clusters from the cell (give preference to meta clusters, if exist), and compute a hyperplane that separates the selected pair with the maximum margin. The vector is:
\begin{equation}
  \begin{aligned}
    \bm{h} = \frac{\bar{\bm{f}}_i - \bar{\bm{f}}_j}{2}.
  \end{aligned}
\end{equation}
This hyperplane can divide the cell into two parts. This process continues until there are no cells containing more than two meta clusters and contain no more than $B$ clusters. The overall division process is listed in Algorithm~\ref{alg:LSH} and \ref{alg:TB_cluster}. Finally, we can obtain the final top-down subgroup labels $\bm{c}^{T}$.

\begin{algorithm}[t]
  \caption{Top\_Down\_Division}
  \label{alg:LSH}
  \begin{algorithmic}[1]
    \Require
    A set of subgroups $\mathcal{C}=\{\mathcal{C}_i\}^{M_c}_{i=1}$;
    The maximum number of samples in one cluster $B$
    \If {$\sum_{i=1}^{M_c} \vert \mathcal{C}_i \vert \geq B \wedge M_c > 1$}
    \If { More than one meta clusters exist in $\mathcal{C}$}
    \State Sample two different meta clusters $\mathcal{C}_i^*, \mathcal{C}_j^*$
    \Else
    \State Sample two different subgroups $\mathcal{C}_i, \mathcal{C}_j$
    \EndIf
    \State Get the hyperplane vector $\bm{h} = \frac{\bar{\bm{f}}_i - \bar{\bm{f}}_j}{2}$ 
    \State $\mathcal{C}^l = \{\mathcal{C}_k: \bar{\bm{f}}_k^{\,\top}\bm{h} \geq 0\}$
    \State $\mathcal{C}^r = \{\mathcal{C}_k: \bar{\bm{f}}_k^{\,\top}\bm{h} < 0\}$
    \State $\mathcal{C}^l_\text{sub}$ = Top\_Down\_Division($\mathcal{C}^l$, $B$)
    \State $\mathcal{C}^r_\text{sub}$ = Top\_Down\_Division($\mathcal{C}^r$, $B$)
    \State $\mathcal{C}_\text{sub} = \{\mathcal{C}^l_\text{sub}, \mathcal{C}^r_\text{sub}\}$
    \Else
    \State $\mathcal{C}_\text{sub} = \{\mathcal{C}\}$
    \EndIf
    \Ensure $\mathcal{C}_\text{sub}$
  \end{algorithmic}
\end{algorithm}
\begin{algorithm}[t]
  \caption{Complementary top-down subgroup generation}
  \label{alg:TB_cluster}
  \begin{algorithmic}[1]
    \Require A set of subgroups $\mathcal{C}=\{\mathcal{C}_i\}^{M_c}_{i=1}$;
    The maximum number of samples in one cluster $B$
    \State Compute and normalize the centroid $\bar{\bm{f}}_i$ of each subgroup
    \State $\mathcal{C}_\text{sub}$ = Top\_Down\_Division($\mathcal{C}$, $B$)
    \Ensure Cluster labels for each sample $\bm{c}^{T} = \{c_i^{T}\}_{i=1}^N$
  \end{algorithmic}
\end{algorithm}

\subsection{Positive Prototype Generation Module}
\begin{figure}[!t]
  \begin{center}
    \includegraphics[width=1.0\linewidth]{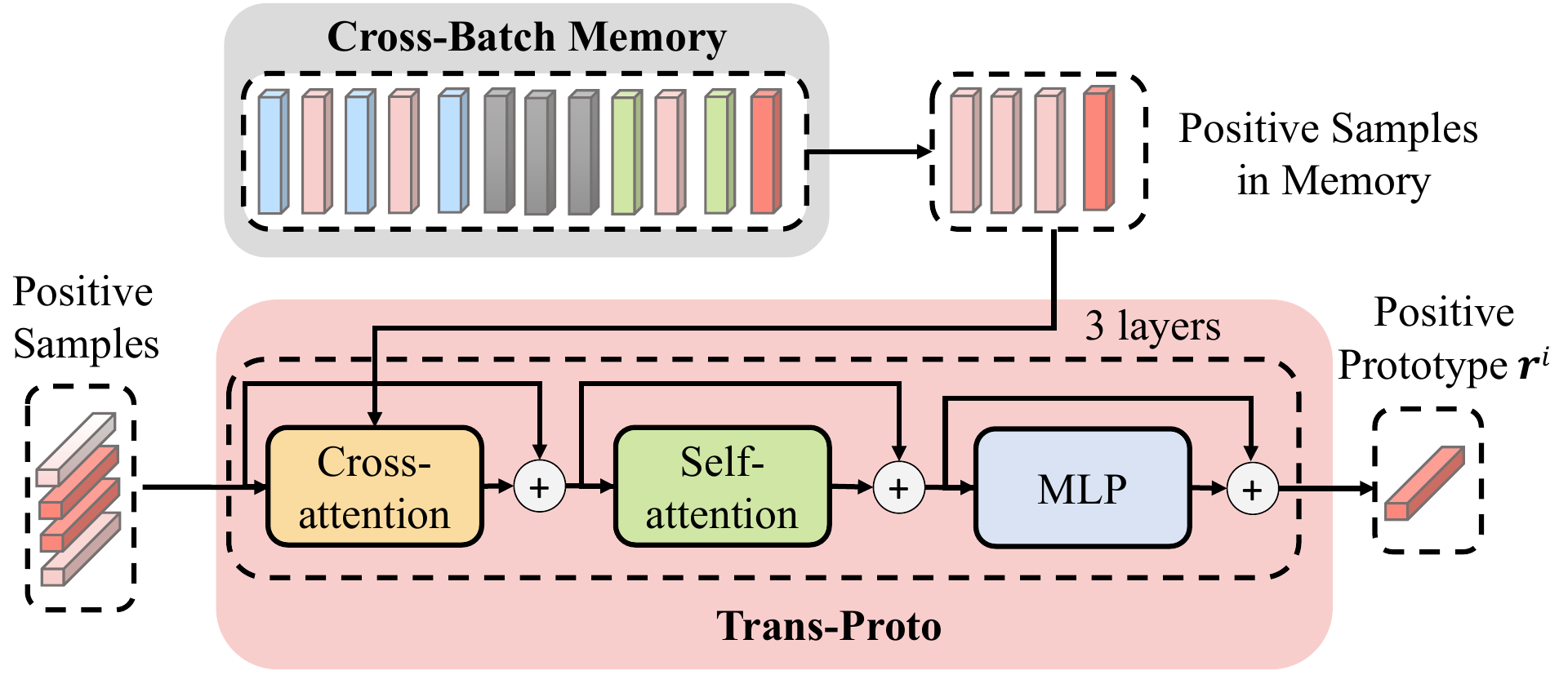}
    \caption{
      Visualization of TransProto PPM. We employ a 3-layer transformer-based model learns to aggregate features of positive samples into the learning prototype $\bm{r}^i$.
    }
    \vspace{-5mm}
    \label{fig:ppm}
  \end{center}
\end{figure}

The subgroup labels $\bm{c}^B, \bm{c}^T$ obtained from SGM provide extra cleaner supervision information. For each noisy sample, compared with directly sampling its positive samples based on the annotated labels, sampling based on subgroup labels is more likely to construct true positive pairs. Therefore, we propose to select extra $K$ positive samples denoted as $\mathcal{P}_i$ based on $\bm{c}^B$ and $\bm{c}^T$ for the $i$-th noisy sample in $\setB_\text{noise}$. To further eliminate the impact of possible noisy samples in $\mathcal{P}_i$ without losing valuable hard positive samples, we propose to aggregate the features of $K$ positive samples into a single reliable prototype $\bm{r}^{i}$. Several prototype aggregation methods are designed as follows:
\begin{itemize}
  \item \textbf{Mean:}
        A straightforward approach is taking the mean of the features of all positive samples corresponding to the $i$-th noisy sample as the positive prototype:
        \begin{equation}
          \begin{aligned}
            \bm{r}_{mean}^i = \frac{1}{K} \sum_{k=1}^{K} f\left(x_k\right),
          \end{aligned}
          \label{eq:mean_proto}
        \end{equation}
        where $x_k \in \mathcal{P}_i$.

  \item \textbf{Max:}
        Instead of treating all positive samples equally, we choose the sample most similar to the noisy sample as the positive prototype:
        \begin{equation}
          \begin{aligned}
            j^{\star}      & = \mathop{argmax}\limits_{j \in [K]} S_{ij}, \\
            \bm{r}_{max}^i & = f(x_{j^{\star}}).
          \end{aligned}
          \label{eq:max_proto}
        \end{equation}

  \item \textbf{SoftMax:}
        In order to better utilize the information of all positive samples, we conduct the aggregation by weighting them based on their correlation. Concatenating the extracted features from $\mathcal{P}_i$ as a feature matrix $\bm{F}_i\in\mathbb{R}^{K\times d}$, then we have
        \begin{equation}
          \begin{aligned}
            \bm{Corr}_i     & = \bm{F}_i \bm{F}_i^{\top} - \bm{I},                                          \\
            \bm{r}_{soft}^i & = \text{softmax}(\frac{1}{K} \bm{Corr}_i \cdot \bm{1})^{\top} \cdot \bm{F}_i,
          \end{aligned}
          \label{eq:soft_protos}
        \end{equation}
        where $\bm{I}$ is the $K\times K$ identity matrix and $\bm{1}$ is a $K\times 1$ all-one vector. Namely, the weight of each sample in $\mathcal{P}_i$ is measured by summing up their correlations to other samples in this set.

  \item \textbf{TransProto:}
        In addition to the fixed aggregation strategies, we introduce a novel learnable mechanism based on a transformer for enhanced flexibility in aggregation. Illustrated in Fig.\ref{fig:ppm}, we concatenate the extracted features of $\mathcal{P}_i$ to create an input sequence for a 3-layer cross-attention-based transformer. It is equipped with the cross-batch memory bank $\mathcal{M}$ to store historical features. Leveraging the historical features of samples within the same subgroup as $\mathcal{P}_i$ as the conditioned feature for cross-attention, we add the transformer's output up and normalize them as the prototype $\bm{r}^i$. Unlike the aforementioned methods that merely assign weights to existing positive samples, TransProto works as a learned mechanism that leverages the powerful information extraction capabilities of cross-attention. This allows it to generate novel prototypes distinct from the original positive feature distribution. As a result, TransProto exhibits the ability to handle more complex noisy label scenarios.
\end{itemize}

The cross-batch memory in Fig.\ref{fig:ppm} is actually an extension of XBM\cite{wang2020cross}, which is maintained as a queue with the current mini-batch enqueued and the oldest mini-batch dequeued. In addition to features which are the only contents in XBM, our cross-batch memory also stores the indicator of being clean samples generated by the PCS module together with the corresponding noisy labels and subgroup labels. These contents will be also used to update the centroids in PCS, generate positive pairs for noisy samples in PPM and provide negative pairs for the memory-bank-based contrastive loss (MCL).

\subsection{Loss Functions}
\noindent\textbf{Loss for $\setB_{\text{clean}}$}:
The traditional pair-wise contrastive loss function computes similarities between all pairs of data samples. The loss function encourages $f(\cdot)$ to assign small distances between samples in the same class and large distances between samples from different classes. More formally, a typical contrastive loss (CL) over $\setB_{\text{clean}}$ is:
\begin{equation}
  \begin{aligned}
    L_{\text{batch}} & = \sum_{\substack{i,j \in \setB_{\text{clean}} \\ y_i \neq y_j}}
    \left[S_{ij}-\lambda\right]_{+}   -
    \sum_{\substack{i,j \in \setB_{\text{clean}}                      \\ y_i=y_j}}
    S_{ij},
  \end{aligned}
\end{equation}
where $\lambda \in[0,1]$ is the predefined margin hyperparameter and $[x]_{+}= \max (x, 0)$. With the cross-batch memory $\mathcal{M}$ same in PCS and PPM that stores the features of data samples $\bm{v}$, we can obtain more positive and negative pairs in the loss, which may reduce the variance in the gradient estimation. The memory-bank-based contrastive loss (MCL)~\cite{wang2020cross} can be written as:

\begin{equation}
  \begin{aligned}
    L_{\text{bank}} = \sum_{i \in \setB_\text{clean}} \bigg[ & \sum_{\substack{j \in \mathcal{M}   \\
    y_i \neq y_j}} [S_\param\left(f(\bm{x}_i), \bm{v}_j\right)-\lambda]_+                          \\
                                                             & - \sum_{\substack{j \in \mathcal{M} \\
          y_i=y_j}} S_\param\left(f(\bm{x}_i), \bm{v}_j\right) \bigg].
  \end{aligned}
\end{equation}
The total loss over clean samples is the sum of the basic batch loss $L_{\text{batch}}$ and the memory bank loss $L_{\text{bank}}$, which is denoted as:
\begin{equation}
  L_{\text{clean}} = L_{\text{batch}} + L_{\text{bank}}.
  \label{eq:loss_clean}
\end{equation}
Other novel DML methods like SoftTriple~\cite{qian2019softtriple} can also be used as the loss function for clean samples.

\noindent\textbf{Loss for $\setB_\text{noise}$}:
We introduce a novel loss function for samples in $\setB_\text{noise}$ based on the prototypes generated in PPM. Denote the feature of $i$-th sample in $\setB_\text{noise}$ as $\bm{z}_i=f_\param(\bm{x}_i)$ and $\bm{r}_i$ is the corresponding positive prototype generated by PPM from $\mathcal{P}_{i}$, the loss function for each noisy sample is defined as:
\begin{equation}
  \begin{aligned}
    L_\text{SGPS}^\text{batch} = - \log\frac{\exp((\bm{z}_i\cdot\bm{r}_{i} -\delta)/ \tau) }{\exp
      ((\bm{z}_i\cdot\bm{r}_{i} -\delta)/\tau)+\sum\limits_{\bm{z}_j \in \mathcal{N}_i} \exp(\bm{z}_i\cdot\bm{z}_j/\tau)}
  \end{aligned}
  \label{eq:loss_ppm_batch}
\end{equation}
where $\tau$ is the temperature hyperparameter, $\delta$ is a margin parameter to control the distance between positive and negative samples. $\mathcal{N}_i$ is the set of negative samples for $i$-th sample, which is defined as:
\begin{equation}
  \begin{aligned}
    \mathcal{N}_i = \{\bm{z}_j | y_j \neq y_i \, \wedge \, c_j^{B} \neq c_i^{B} \, \wedge \, c_j^{T} \neq c_i^{T}, \, j \in \setB\}.
  \end{aligned}
  \label{eq:negative_set}
\end{equation}
Similar to $L_\text{bank}$, we employ the feature memory bank $\mathcal{M}$ to further increase the negative pairs in the loss. The memory bank based loss can be defined as:
\begin{equation}
  \begin{aligned}
    L_\text{SGPS}^\text{bank} = - \log\frac{\exp((\bm{z}_i\cdot\bm{r}_{i} -\delta ) / \tau)}{\exp
      ((\bm{z}_i\cdot\bm{r}_{i} -\delta)/\tau)+\sum\limits_{\bm{z}_j \in \mathcal{N}_i^\text{bank}} \exp(\bm{z}_i\cdot\bm{z}_j/\tau)},
  \end{aligned}
  \label{eq:loss_ppm_bank}
\end{equation}
where $\mathcal{N}_i^\text{bank}$ is the set of negative samples for $i$-th sample defined similar to Eq.\eqref{eq:negative_set} except that $j$ is from $\mathcal{M}$. Therefore, the loss over the noisy subset is:
\begin{equation}
  L_{\text{noise}} = \gamma_{1} \cdot L_{\text{SGPS}}^{\text{batch}} + \gamma_{2} \cdot L_{\text{SGPS}}^{\text{bank}},
  \label{eq:loss_noise}
\end{equation}
where $\gamma_{1}$ and $\gamma_{2}$ is loss weight to balance the noisy sample loss on batch and bank.

\noindent\textbf{Overall.} Putting all together, the total objective of SGPS is:
\begin{equation}
  L_\text{all} = L_\text{clean} + L_\text{noise}.
  \label{eq:loss_all}
\end{equation}

\section{Experiments}
To evaluate the effectiveness of the proposed approach on noisy DML tasks, we compare it against 13 baseline methods on both synthetic and real-world image retrieval benchmarks. Further, we conduct experiments on large-scale face recognition tasks to demonstrate the generalization of our approach.

\subsection{Datasets}
The evaluation is conducted on seven image retrieval or face recognition benchmark datasets, including:

The evaluation is conducted on seven image retrieval or face recognition benchmark datasets, including:
\begin{itemize}
  \item \textbf{CARS}~\cite{CARS196} contains 16,185 images of 196 different car models. We use the first 98 models for training, the rest for testing.
  \item \textbf{CUB}~\cite{CUB} contains 11,788 images of 200 different bird species. We use the first 100 species for training and the rest for testing.
  \item \textbf{Stanford Online Products (SOP)}~\cite{SOP} contains 59,551 images of 11,318 furniture items on eBay. We use 59,551 images in all classes for training and the rest for testing.
  \item \textbf{Food-101N}~\cite{FOOD101N} is a real-world noisy dataset that contains 310,009 images of food recipes in 101 classes. It has the same 101 classes as Food-101~\cite{FOOD101} (which is considered a clean dataset). We use 144,086 images in the first 50 classes (in alphabetical order) as the training set, and the remaining 51 classes in Food-101 as the test set which contains 51,000 images.
  \item \textbf{CARS-98N}~\cite{PRISM2021} is a real-world noisy dataset that contains 9,558 images for 98 car models. The noisy images often contain the interior of the car, car parts, or images of other car models. The CARS-98N is only used for training, and the test set of CARS is used for performance evaluation.
  \item  \textbf{MS1MV0}~\cite{MS1MV0} contains 10M face images of 100K identities collected from the search engine based on a name list, in which there is around 50\% noise.
  \item  \textbf{MS1MV2}~\cite{deng2019arcface} contains 5.8M face images of 85K identities, which is the cleaned version of MS1MV0~\cite{MS1MV0} by a semi-automatic pipeline.
  \item \textbf{Clothing1M}~\cite{Cloth1M} { is a large-scale real-world dataset with noisy labels. It contains 1M images from 14 different cloth-related classes. We use the 1M noisy images for training and same testset as classification with 10,525 clean images for testing.}
\end{itemize}

We employ two types of synthetic label noise: 1) symmetric noise~\cite{van2015learning} and 2) small cluster noise~\cite{PRISM2021}. Symmetric noise is widely utilized to assess the robustness of classification models. Under this model, a predefined portion of data from each ground truth class is assigned to all other classes with equal probability, irrespective of the similarity between data samples. The number of classes remains unchanged after applying this noise synthesis. On the other hand, small cluster noise emulates naturally occurring label noise by iteratively flipping labels. In each iteration, images are clustered from a randomly selected ground-truth class into numerous small clusters, and each cluster is then merged into another randomly selected ground-truth class. This process creates an open-set noisy label scenario~\cite{wang2018iterative}, where some images do not belong to any other existing class in the training set.

\subsection{Baselines}
We compare SGPS against 19 baseline approaches on image retrieval tasks, including
(1) noise-resistant classification methods: Co-teaching~\cite{han2018co}, Co-teaching+~\cite{coteaching+}, Co-teaching with Temperature~\cite{nSoftmax}, and F-correction~\cite{F-correction}, {Unicon~\cite{karim2022unicon}, DivideMix~\cite{li2020dividemix}, LRA-Diffusion\cite{chen2024label} and {LabelWave} \cite{yuanearly}.}
(2) DML methods with proxy-based losses: SoftTriple~\cite{qian2019softtriple}, FastAp~\cite{FastAp}, nSoftmax~\cite{nSoftmax} and proxyNCA~\cite{proxyNCA};
(3) DML methods with pair-wise losses: MS loss~\cite{wang2019multi}, circle loss~\cite{sun2020circle}, contrastive loss~\cite{chopra2005learning}, memory contrastive loss (MCL)~\cite{wang2020cross},
{SupCon~\cite{khosla2020supervised}, Roadmap~\cite{ramzi2021robust}, Contextual~\cite{liao2023supervised} } and PRISM~\cite{PRISM2021}.
Among noise-resistant classification baselines, F-correction assumes closed-world noise and can only be used under symmetric noise. We train the classification baselines using the cross-entropy loss and use the l2-normalized features from the penultimate layer when retrieving images during inference. {On the other hand, to demonstrate the applicability of our framework, we instantiated our SGPS on four DML losses: MCL, SupCon, Roadmap and Contextual.}

For all experiments on image retrieval, a consistent batch size of 64 is set and conducted on one NVIDIA RTX 3090 GPU. During training, the input images are first resized to 256$\times$256, then randomly cropped to 224$\times$224. A horizontal flip is performed on the training data with a probability of 0.5. The validation and testing images are resized to 224$\times$224 without data augmentation.  Following~\cite{wang2020cross}, when comparing performance on CARS, CUB and CARS-98N, we use BN-inception~\cite{ioffe2015batch} as the backbone model. The dimension of the output feature is set as 512. For SOP and Food-101N datasets, we use ResNet-50~\cite{he2016deep} with a 128-dimensional output.
In terms of evaluation metrics, we base our assessment on the ranked list of nearest neighbors for test images. Specifically, Precision@1 (P@1) and Mean Average Precision@R (MAP@R)~\cite{musgrave2020metric} are adapt as our evaluation metrics.

For the large-scale metric learning task, i.e., face recognition, we follow ArcFace~\cite{deng2019arcface} to get the aligned face crops and resize them into (112$\times$112). Then, a ResNet-like~\cite{he2016deep} network R50 is used to extract representation and returns a 512-D embedding for each image. The experiments of face recognition are implemented by PaddlePaddle~\cite{ma2019paddlepaddle} and trained on 8 NVIDIA 3090 GPUs with a total batch size of 512. For proxy-based methods, the class center is a learnable vector, the same as the classifier. We set the learning rate as 0.1 and use 64 and 0.4 as the scale and margin for other proxy-based methods. For pair-wise methods, including ours, we set the learning rate as 0.06 at the start of training and downscale it by 0.1 at 4th, 8th and 9th epoch. The training process ends at the 10th epoch. The weight decay is set to 1e-4, and the momentum of the SGD optimizer is 0.9. We adopt the loss function in DCQ~\cite{li2021dynamic} as $L_\text{clean}^\text{face}$:
\begin{equation}
  \begin{aligned}
    {L}_\text{clean}^\text{face}=-\ln \frac{e^{s\left(\cos \left(\theta_y\right)-m\right)}}{e^{s\left(\cos \left(\theta_y\right)-m\right)}+\sum_{j \in [M] / \{y\}} e^{s \cos \left(\theta_j\right)}},
  \end{aligned}
  \label{eq:DCQ}
\end{equation}
where $\cos(\theta_y)$ is the cosine similarity between the feature vector and the corresponding class center (dynamically generated from the class queue~\cite{li2021dynamic}). The scale $s$ and margin $m$ in the loss are set to 50 and 0.3, respectively.

\begin{table*}[t]
  \setlength\tabcolsep{1.5mm}
  \centering
  \caption{ Precision@1 (\%) on CARS, SOP, and CUB datasets with symmetric label noise. \textcolor{Red}{\textbf{BEST}} and \textcolor{Blue}{\textit{\textbf{SECOND}}} best values are both highlighted in bold. SGPS-MCL$\space^{\star}$ indicates we perform the posterior data clean strategy.
  }
  \label{tab:symmetric}%
\begin{tabular}{lccccccccccccccccc}
    \toprule
    \multicolumn{1}{r}{}                                          & \multicolumn{5}{c}{CARS}                  &                                           & \multicolumn{5}{c}{SOP}                   &                                           & \multicolumn{5}{c}{CUB}                                                                                                                                                                                                                                                                                                                                                                                                                                                                                 \\
    \cmidrule{2-6} \cmidrule{8-12} \cmidrule{14-18}    Noisy Rate & 10\%                                      & 20\%                                      & 50\%                                      & 70\%                                      & 90\%                                      &  & 10\%                                      & 20\%                                      & 50\%                                      & 70\%                                      & 90\%                                      &  & 10\%                                      & 20\%                                      & 50\%                                      & 70\%                                      & 90\%                                      \\
\bottomrule
  \end{tabular}%
\end{table*}%

\begin{table*}[t]
  \setlength\tabcolsep{3mm}
  \centering
  \caption{Precision@1 (\%) on CARS, SOP, and CUB with Small Cluster label noise.}
  \label{tab:smallcluster}%
  \begin{tabular}{lcccccccccccccc}
    \toprule
    \multicolumn{1}{r}{}                                         & \multicolumn{3}{c}{CARS}                  &                                           & \multicolumn{3}{c}{SOP}                   &       & \multicolumn{3}{c}{CUB}                                                                                                                                                                                                                                                                                                               \\
    \cmidrule{2-4} \cmidrule{6-8} \cmidrule{10-12}    Noisy Rate & 25\%                                      & 50\%                                      & 75\%                                      &       & 25\%                                      & 50\%                                      & 75\%                                                     &                                 & 25\%                                      & 50\%                                                     & 75\%                                      \\
    \midrule
    \multicolumn{8}{l}{\textit{Algorithms for image classification under label noise}}                                                                                                                                                                                                                                                                                                                                                                                                                                                               \\
\bottomrule
  \end{tabular}%
\end{table*}%

\begin{table}[t]
  \setlength\tabcolsep{0.74mm}
  \centering
  \caption{Precision@1 (\%) and Mean Average Precision@R
    (\%) on CARS-98N, Food-101N and Clothing1M.}
  \label{tab:realworld}%

  \begin{tabular}{lcccccccc}
    \toprule
    \multicolumn{1}{r}{}                                                   & \multicolumn{2}{c}{CARS-98N}              &                                           & \multicolumn{2}{c}{ Food-101N} &                                           & \multicolumn{2}{c}{ Clothing1M}                                                                                                      \\
    \cmidrule{2-3}\cmidrule{5-6}\cmidrule{8-9} \multicolumn{1}{r}{} Metric & \multicolumn{1}{c}{P@1}                   & \multicolumn{1}{c}{MAP@R}                 &                                & \multicolumn{1}{c}{P@1}                   & \multicolumn{1}{c}{MAP@R}                 &  & \multicolumn{1}{c}{P@1}                   & \multicolumn{1}{c}{MAP@R}                 \\
    \midrule
    \multicolumn{6}{l}{\textit{Algorithms for image classification under label noise}}                                                                                                                                                                                                                                                                                                 \\
    Co-teaching~\cite{han2018co}                                           & 58.74                                     & 9.10                                      &                                & 59.08                                     & 14.66                                     &  & 70.41                                     & 46.43                                     \\
    Co-teaching+~\cite{coteaching+}                                        & 56.66                                     & 8.40                                      &                                & 57.59                                     & 14.72                                     &  & 70.53                                     & 47.89                                     \\
\end{tabular}%
\end{table}%

\begin{table*}[t]
  \centering
  \caption{Experiments of different settings on MS1MV0
    dataset comparing with state-of-the-art methods.
    The number in \textcolor{Red}{Red} denote the best result.
    \textcolor{Red}{\textbf{BEST}} and \textcolor{Blue}{\textit{\textbf{SECOND}}} best values
    in algorithms without classifier are both highlighted in bold.
    We adopt the 1:1 verification TAR (@FAR=1e-3,1e-4,1e-5) on
    the IJB-B and IJB-C dataset as the evaluation metric.
  }
  \label{tab:face}%
  \begin{tabular}{l|l|rrr|rrr|rrr}
    \toprule
    Method                                                  & Dataset & LFW                                       & CFP-FP                                    & AgeDB-30                                  & \multicolumn{3}{c}{IJB-B }                & \multicolumn{3}{c}{IJB-C}                                                                                                                                                                                                 \\
                                                            &         &                                           &                                           &                                           & 1e-03                                     & 1e-04                                     & 1e-05                                     & 1e-03                                     & 1e-04                                     & 1e-05                                     \\
    \midrule
    \multicolumn{11}{l}{\textit{Algorithms with classifier}}                                                                                                                                                                                                                                                                                                                                                                                                                      \\
    ArcFace~\cite{deng2019arcface}                          & MS1Mv0  & 99.75                                     & 97.17                                     & 97.26                                     & 93.27                                     & 87.87                                     & 74.74                                     & 94.59                                     & 90.27                                     & 81.11                                     \\
\bottomrule
  \end{tabular}%
\end{table*}%

\subsection{Results}

\noindent\textbf{Symmetric Label Noise.}
Table \ref{tab:symmetric} shows the evaluation results on CARS, SOP, and CUB under symmetric label noise. Our methods achieve the highest performance in all the cases. As the noise rate increases, all approaches experience a decrease in Precision@1 scores. {Notably, our SGPS-based models instantiated on different DML losses are more stable towards the noise rate. In the case of smaller datasets like CARS and CUB, SGPS-based models exhibit a decrease of less than 6\% in P@1 as the noise level increases from 10\% to 50\%.
    In contrast, most competitors suffer from a performance drop of more than 10\%.
    For example, the recent noisy label learning method LRA-Diffusion show promising results on CARS, but drop significantly in SOP where the class distribution is imbalanced and the number of images per class decreases.
    We can see that DivideMix and UNICON both perform not very well, as they are designed for image classification tasks, the guessing label strategy may not be suitable for DML tasks.
  }
  {On the other hand, Roadmap exhibits relatively stable performance among DML methods. The reason might be that the calibration loss in Roadmap enforces the score of the negative pairs to be smaller than $\beta$, which alleviates the impact of noisy labels to some extent.}
PRISM effectively improves the ratio of correct pairs by preserving clean samples with high confidence.
Nevertheless, discarding noisy samples could impair overall performance. Our framework takes a step further by effectively utilizing noisy samples, leading to superior performance compared to all other methods.
When the noise rate increases, SGPS suffers less performance degradation compared to the competitors, where SGPS-Contextual under 70\% noise rate even outperforms most competitors under 50\% noise rate.

\noindent\textbf{Small Cluster Label Noise.}
Table \ref{tab:smallcluster} reports the Precision@1 scores on datasets with small-cluster label noise. The results are similar to the case of symmetric noise. Contextual achieves a very high performance in the low noise rate setting.
When trained on CARS on 25\% small cluster noise, Contextual achieves a higher performance than PRISM-MCL. However, in the case of high noise rates, it drops significantly. Our SGPS-Contextual outperforms SoftTriple and PRISM-MCL by a significant margin across all datasets. This success further demonstrates the importance of utilizing the noisy samples rather than discarding them.

\noindent\textbf{Real-world Noise.}
Table \ref{tab:realworld} displays the performance on three datasets with real-world label noise, CARS98N and Food-101N~\cite{FOOD101N} and Clothing1M \cite{Cloth1M}. Unlike the previous datasets, these three datasets contain a significant number of out-of-distribution (OOD) samples. The actual number of categories and the total noise ratio are both unknown.
It is easy to see that the multiProxy-based method SoftTriple is more capable for such a phenomenon, as it can assign multiple centers even for mislabeled training samples.
  {Roadmap also shows good performance as its extra constraints can help to alleviate the influence of a large amount of uncertain negative pairs.}
In contrast, PRISM-MCL and XBM-MCL can not handle this situation well. To handle such a complicated noise situation, we set $\tau_{k} $ to be twice the number of categories to enhance the accuracy of positive pairs selected by SGM.
  {
    Considering the test set of Clothing1M shares the same 14 categories as its training set, the performance difference among different methods might be not very large.
    Nonetheless, SGPS-Roadmap achieves the best performance on all three datasets, and the proposed SGPS-based framework could improve the performance of its base model (SupCon, MCL, Contextual, and Roadmap).} This again demonstrates the strength of integrating advanced DML methods with our SGPS framework in real-world cases.


\noindent\textbf{Large-scale Noisy Dataset.}
Tab.\ref{tab:face} shows the results on the large-scale noisy face recognition dataset MS1MV0~\cite{MS1MV0}. The training of ArcFace~\cite{deng2019arcface} and CosFace~\cite{wang2018cosface} on MS1MV0 is hindered by the gradient conflict arising from massive fine-grained intra-class noise and inter-class noise, resulting in limited performance. Approaches like NT~\cite{hu2019noiseNT}, NR ~\cite{zhong2019unequalNR}, and Co-mining~\cite{wang2019comining} assign different weights or margins for clean and noisy samples, effectively improving performance by fully leveraging clean data.  Sub-center~\cite{deng2020subcenter} introduces multiple class centroids in the classifier to address intra-class conflict, while SKH~\cite{liu2021switchable} goes a step further by assigning multiple switchable classification layers to handle inter-class conflict. Although algorithms with classifiers can effectively alleviate noise influence compared to those without classifiers, they are time-consuming and resource-intensive when training on large datasets with millions of identities. Recently, methods like DCQ~\cite{li2021dynamic} and FFC~\cite{wang2022efficient} aim to mitigate time and GPU memory costs by constructing dynamic class pools as a substitute for classifiers. They can achieve comparable performance with classifier-based methods when training on clean data. However, it is noteworthy that classifier-free methods tend to encounter challenges in handling extensive noise.  DCQ~\cite{li2021dynamic}, the SOTA classifier-free FR method, achieves only 53.86\% on the challenging IJB-C(TAR@FAR=1e-5), resulting in a performance drop of more than 40\% than training on cleaner data MS1MV2~\cite{deng2019arcface}. We train DCQ with SGPS on MS1MV0, resulting in a substantial improvement of recognition accuracy. Moreover, our method obtains the performance of 92.91\% on IJB-C@1e-5, surpassing DCQ trained with clean data by a notable margin. This improvement can be attributed to our method's utilization of noisy data in MS1MV0, which is discarded in MS1MV2. Comparing our method with classifier-based methods proposed for noise-robust training, e.g., SKH, our method achieves comparable performance, but with much less hardware cost when training on large-scale datasets, which is further discussed in Fig.~\ref{fig:memory}.

\subsection{Ablation Studies}
\begin{figure}[t]
  \centering

  \subfigure[Ablation of SGM]{
    \includegraphics[width=0.48\linewidth]{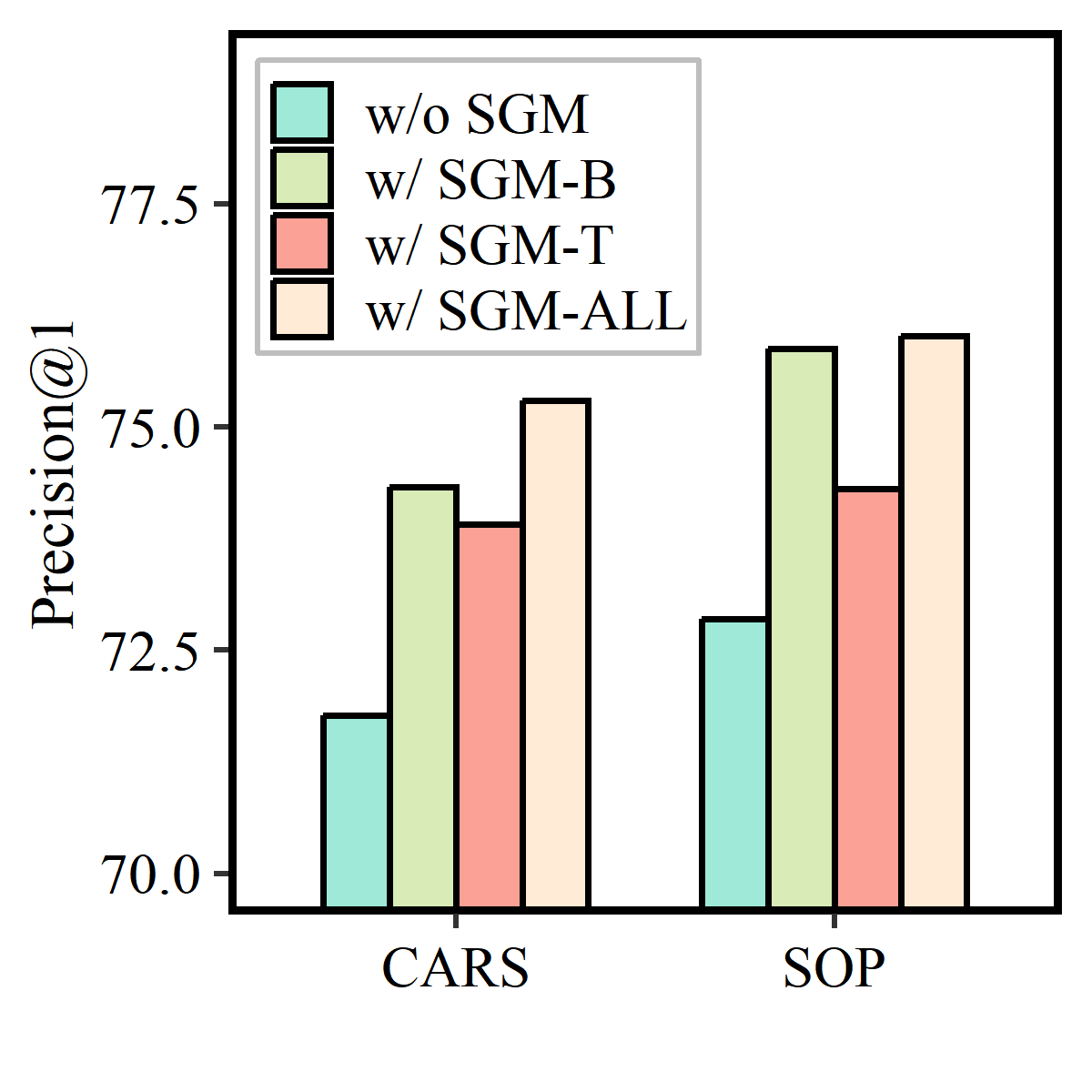}
    \label{fig:ab_sgm}
  }
  \hspace{-0.5cm}
  \subfigure[Ablation of PPM]{
    \includegraphics[width=0.48\linewidth]{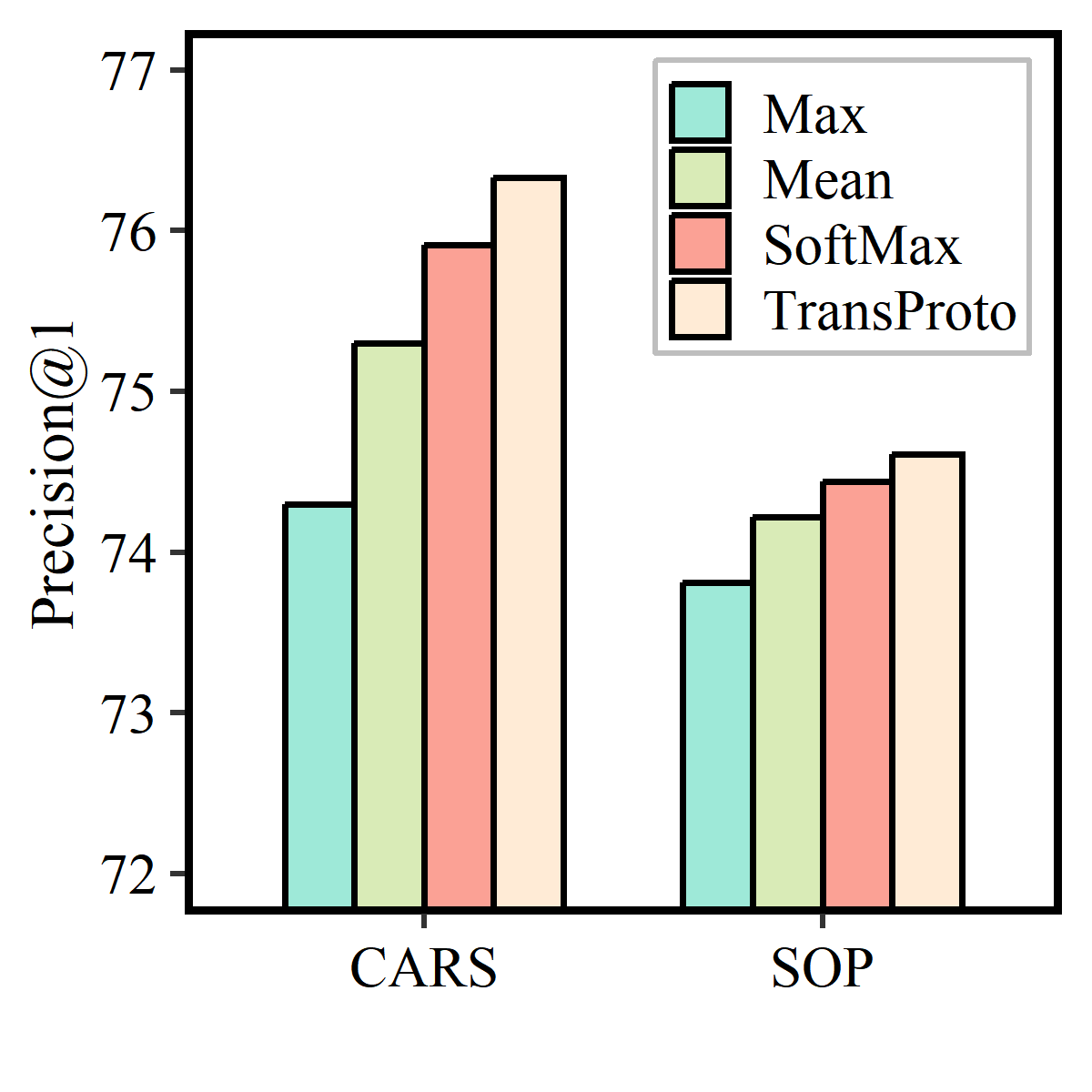}
    \label{fig:ab_ppm}
  }
  \caption{
    Ablation of SGM and PPM on CARS and SOP with 50\% symmetric noise.
  }
  \label{fig:ab}
\end{figure}

\noindent\textbf{Effectiveness of GSM.}
To demonstrate the effectiveness of our proposed SGM, we decouple subgroup modules in SGM to ablate their effectiveness on CARS and SOP with 50\% symmetric noise. As shown in Fig.\ref{fig:ab_sgm}, selecting positive samples from both $\bm{c}^{B}$ and $\bm{c}^{T}$ contributes to the performance enhancement. Notably, when selecting samples from $\bm{c}^{B}$ (SGM-B) a superior Precision@1 is achieved compared to selecting samples from $\bm{c}^{T}$ (SGM-T). This is mainly because $\bm{c}^{B}$ well utilizes the prior knowledge from feature distribution. Nevertheless, $\bm{c}^{T}$ can also perform as a complementary part for $\bm{c}^{B}$ to further improve the performance.

\noindent\textbf{Effectiveness of PPM.}
We also present the results with different aggregation strategies in PPM on CARS and SOP with 50\% symmetric noise in Fig.\ref{fig:ab_ppm}. Notably, weighted-based aggregating methods achieve better P@1 than simply selecting the most similar positive sample, as Max always selects easy positive pairs, leading to rapid overfitting. The performance of TransProto surpasses that of Mean and SoftMax, suggesting the transformer block can discover potentially more effective learning directions when conditioned on the positive pairs in the memory bank.

\begin{table}[t]
  \centering
  \setlength\tabcolsep{2.1mm}
  \caption{Sensitivity of the hyperparameters on CARS19 with 50\% symmetric noise. $\tau$ and $\delta$ are defined in Eq.\eqref{eq:loss_ppm_batch}, while $\gamma_{1}$ and $\gamma_{2}$ are in Eq.\eqref{eq:loss_noise}.}
  \label{tab:ab_hyperparameter}%
  \begin{tabular}{cc|cc|cc}
    \toprule
    $\tau$ & $\delta$ & $\gamma_{1}$ & $\gamma_{2}$ & Precision@1(\%) & MAP@R(\%)      \\
    \midrule
    0.02   & 0.1      & 1.0          & 0.1          & \textbf{76.33}  & \textbf{20.90} \\
    \midrule
    1.0    & 0.1      & 1.0          & 0.1          & 73.05           & 18.56          \\
    0.1    & 0.1      & 1.0          & 0.1          & 74.99           & 19.63          \\
    \midrule
    0.02   & 0.0      & 1.0          & 0.1          & 74.79           & 19.80          \\
    0.02   & 0.3      & 1.0          & 0.1          & 73.16           & 18.56          \\
    \midrule
    0.02   & 0.1      & 2.0          & 0.1          & 75.41           & 19.92          \\
    0.02   & 0.1      & 5.0          & 0.1          & 73.22           & 18.61          \\
    \midrule
    0.02   & 0.1      & 1.0          & 0.5          & 75.56           & 20.60          \\
    0.02   & 0.1      & 1.0          & 1.0          & 74.20           & 19.27          \\
    \bottomrule
  \end{tabular}%
\end{table}%

\noindent\textbf{Sensitivity of hyperparameters.}
The hyperparameter $\tau$ and $\delta$ in Eq.\ref{eq:loss_ppm_batch} and Eq.\ref{eq:loss_ppm_bank} and loss weight $\gamma_1$ and $\gamma_2$ in Eq.\ref{eq:loss_noise} are important in our method. We demonstrate the influence of the hyperparameters in Tab.\ref{tab:ab_hyperparameter}. We can observe that the smaller $\tau$ can achieve better performance, mainly because the smaller $\tau$ will give more weight to the hard negative samples. However, too small $\tau$ will lead to significant fluctuations in loss, resulting in training failures for the model. Increasing $\delta$ obviously improves the performance, However, if $\delta$ becomes excessively large, it can lead to rapid overfitting of $L_\text{SGPS}$, which hampers the learning process for clean samples.

\vspace{-0.4cm}

\begin{figure}[hbt]
  \centering
  \subfigure[CARS]{
    \includegraphics[width=0.45\linewidth]{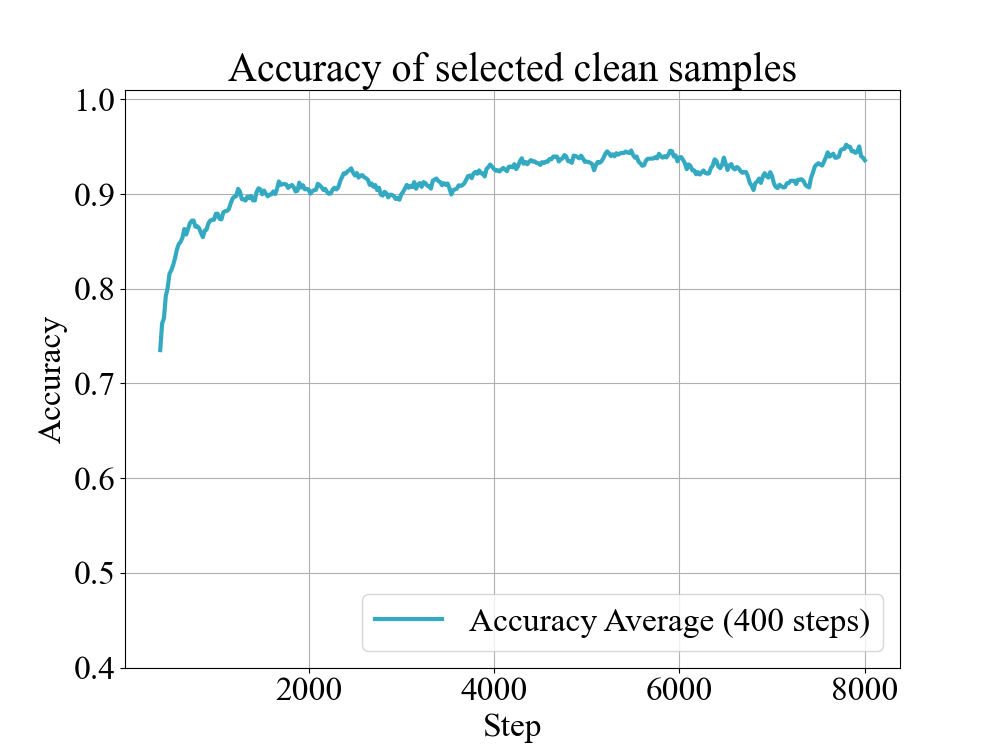}
    \label{pura:cars}
  }
  \subfigure[SOP]{
    \includegraphics[width=0.45\linewidth]{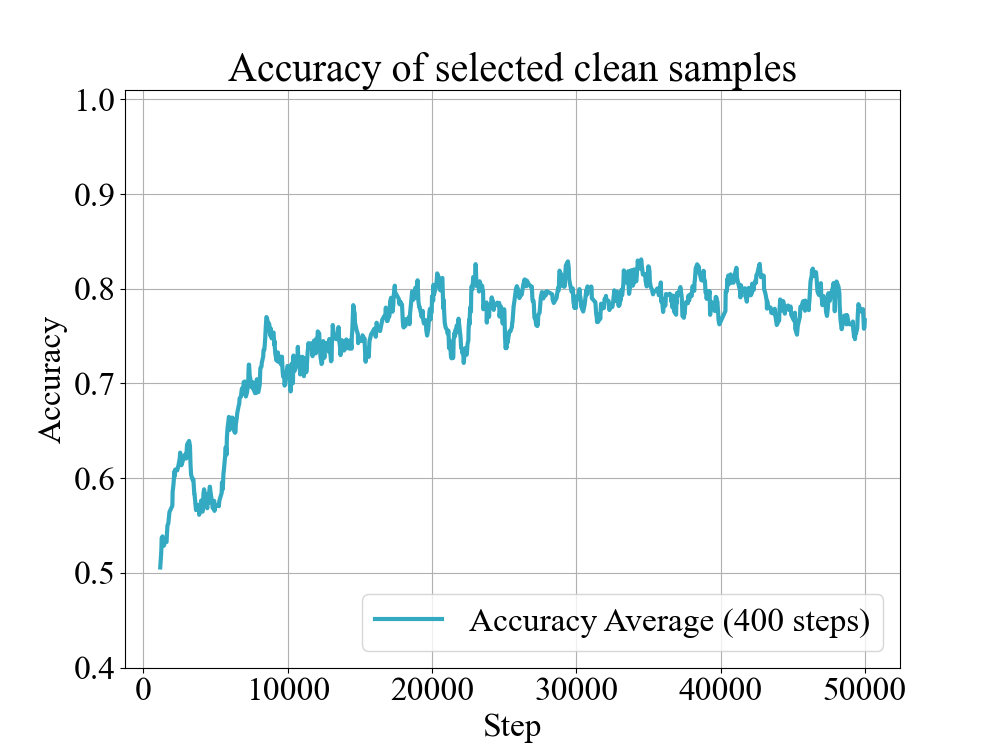}
    \label{pura:sop}
  }
  \caption{Quantitative accuracy of clean sample selection on CARS and SOP with 50\% symmetric noise.}
  \label{fig:filter_acc}
\end{figure}

\begin{figure}[hbt]
  \centering
  \subfigure[CARS-Vis]{
    \includegraphics[width=0.45\linewidth]{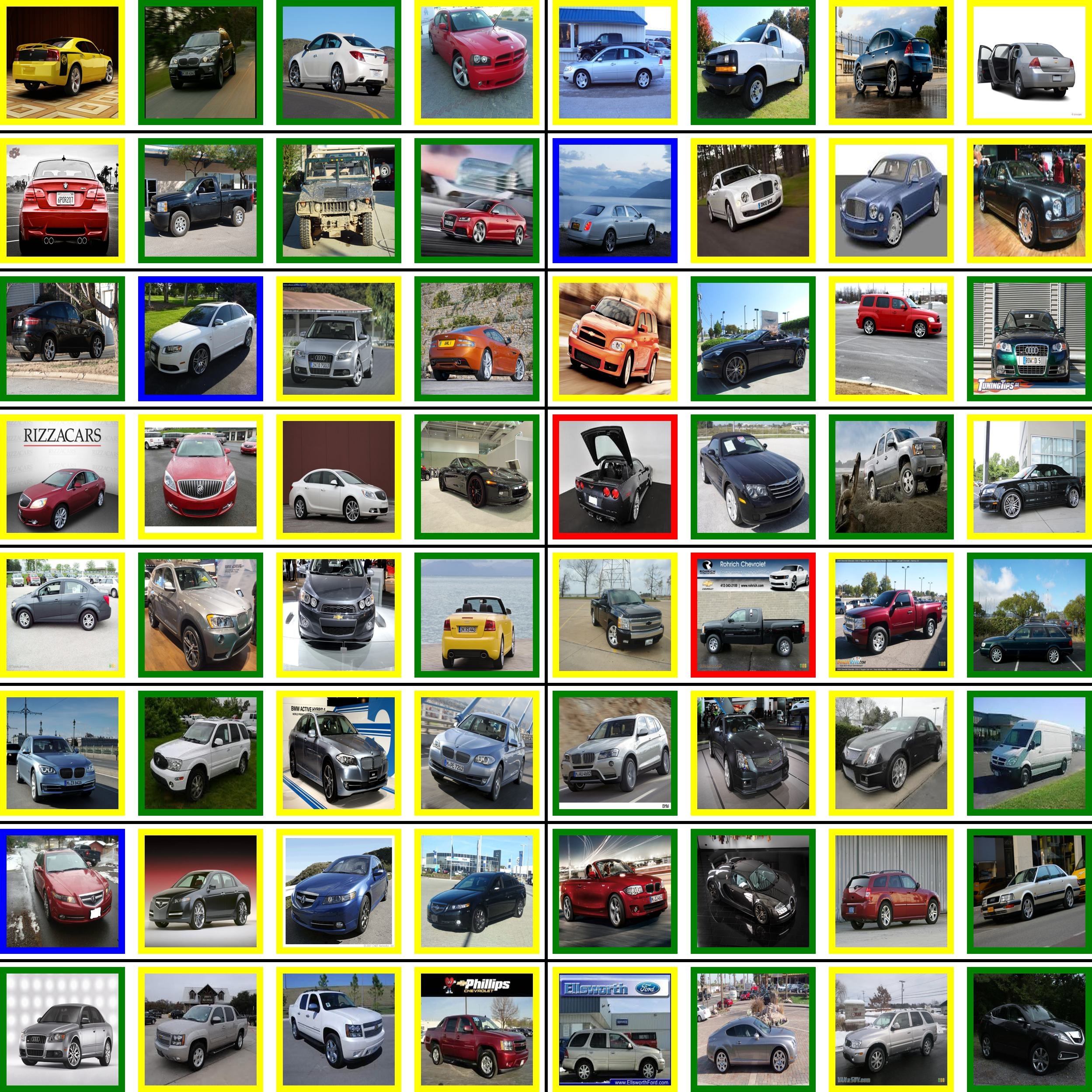}
    \label{pura:carsvis}
  }
  \subfigure[SOP-Vis]{
    \includegraphics[width=0.45\linewidth]{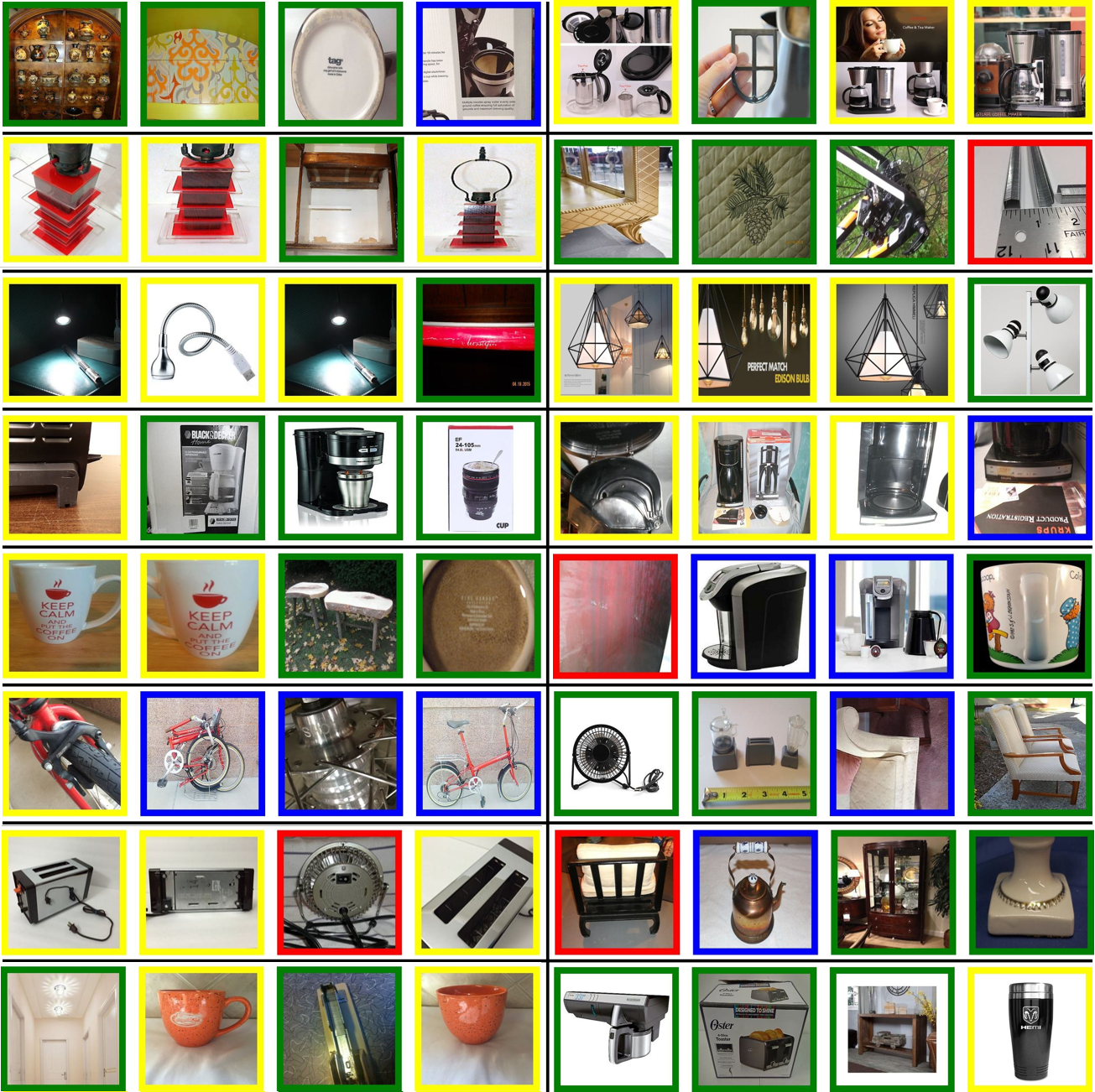}
    \label{pura:sopvis}
  }
  \caption{Qualitative visualization of the detailed selection results in a training batch. \textbf{Yellow} and \textbf{green} boxes represent the selected true clean and discarded true noisy samples, respectively. \textbf{Blue} and \textbf{red} boxes represent the clean sample discarded and the noisy sample selected, respectively. Samples with the same noisy label are separated by \textbf{solid lines.}}
  \label{fig:filter_vis}
\end{figure}

\subsection{Other Analysis\label{sec_analysis}}
\noindent\textbf{Quantitative and qualitative analysis of PCS.}
We present the quantitative accuracy of clean sample selection on CARS and SOP datasets with 50\% symmetric noise in Figure~\ref{fig:filter_acc}.
We can observe the selection accuracy is increasing during training and can achieve more than 90\% on CARS and nearly 80\% on SOP, both much higher than the noise rate (50\%).
Note that the lower accuracy on SOP mainly arises from the fact that it has a notably lower average number of images per class (5.26) than CARS (82.18). Nevertheless, a selection accuracy of 80\% could be sufficiently helpful. To better illustrate the selection results, we also provide the qualitative visualization in a training batch with a size of 64,  in Figure~\ref{pura:carsvis} and Figure~\ref{pura:sopvis}.
It can be observed that only a few noisy samples are mistakenly considered as clean ({\textbf{red boxes}}), while most noisy samples ({\textbf{green boxes}}) are accurately identified and not used in $L_{clean}$. These misselected noisy samples and the corresponding within-class clean samples are semantically similar. For example, in the 5th row of Figure\ref{pura:carsvis}, the misselected noisy sample (red box) is similar to the clean samples (yellow boxes, pickup trucks). The unselected clean samples (blue boxes) are often hard samples within this category.

\noindent\textbf{Posterior Data Clean and Training Strategy.}
We also design a posterior data clean and training strategy to further improve the performance, which contains three stages. First, we train SGPS-MCL with the noisy training set. Once the training process is converged, a DML model with a discriminative feature space is acquired, which can be denoted as a stage-1 model. Then we use stage-1 model to generate pseudo labels, i.e., $\bm{c}^{B}$, for the training set. Second, we train the stage-2 model by the pseudo label without SGPS but with an early stopping PRISM. The stage-2 model can achieve better performance than the stage-1 model by setting a smaller noise rate. In the third step, we train the stage-3 model with the original noisy labels with the stage-2 model as pretrained model and use the extracted feature to initialize the feature of SGM. The related experiments with “\textbf{SGPS-MCL}$\space^{\star}$ ” in Tab.\ref{tab:symmetric} demonstrate the results. Such a strategy can further improve the performance of SGPS-MCL by a large margin, where it outperforms SGPS-MCL by more than 1.2\% on CARS, SOP, and CUB with 50\% symmetric noise. With the proposed clean strategy, we can catch up and even achieve better performance than models trained by clean labels when the noise rate is not very large.

\begin{figure}
  \centering
  \subfigure[Change of Precision]{
    \includegraphics[width=0.46\linewidth]{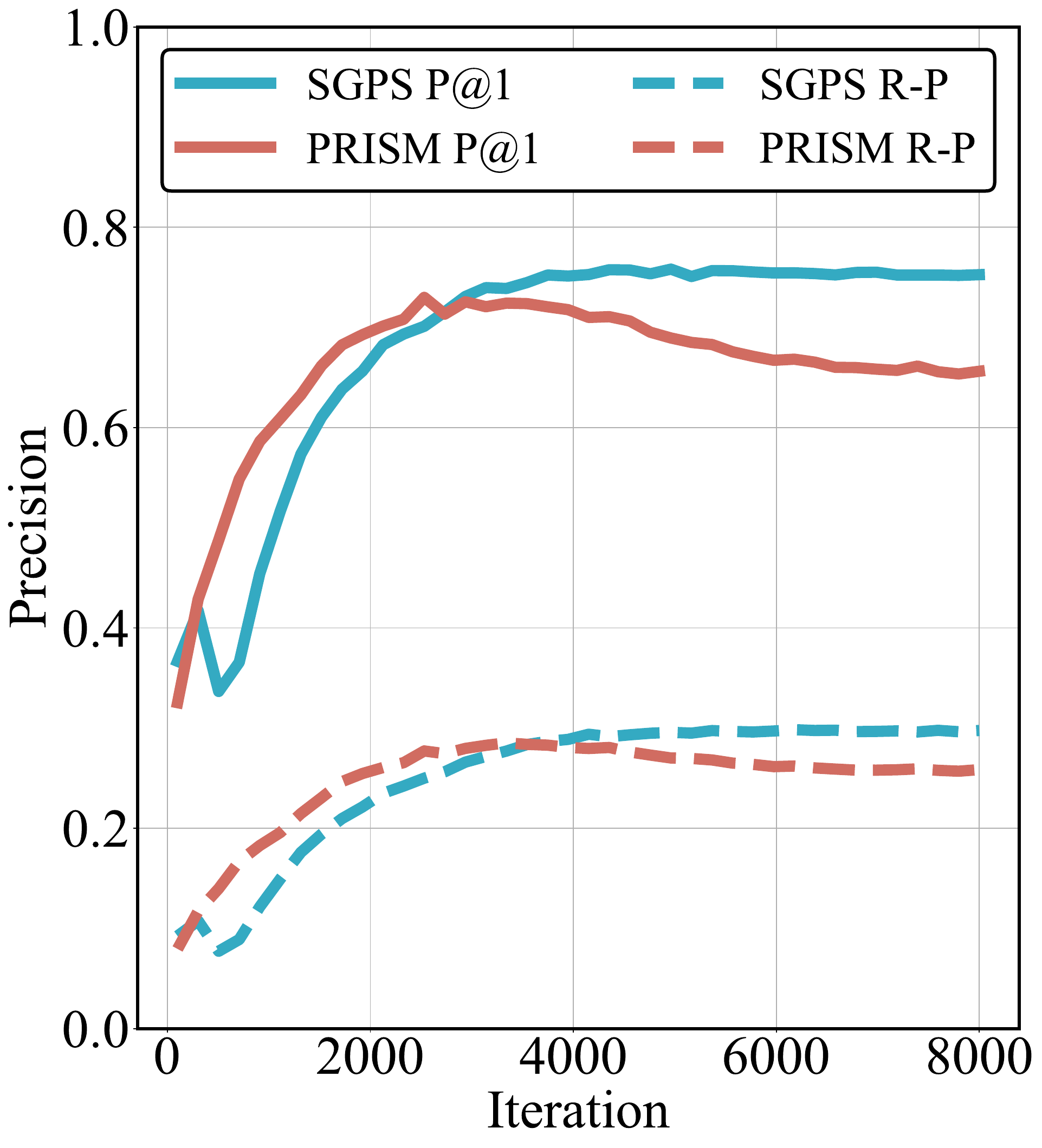}
    \label{fig:vis_precision}
  }
  \subfigure[Change of MAP]{
    \includegraphics[width=0.46\linewidth]{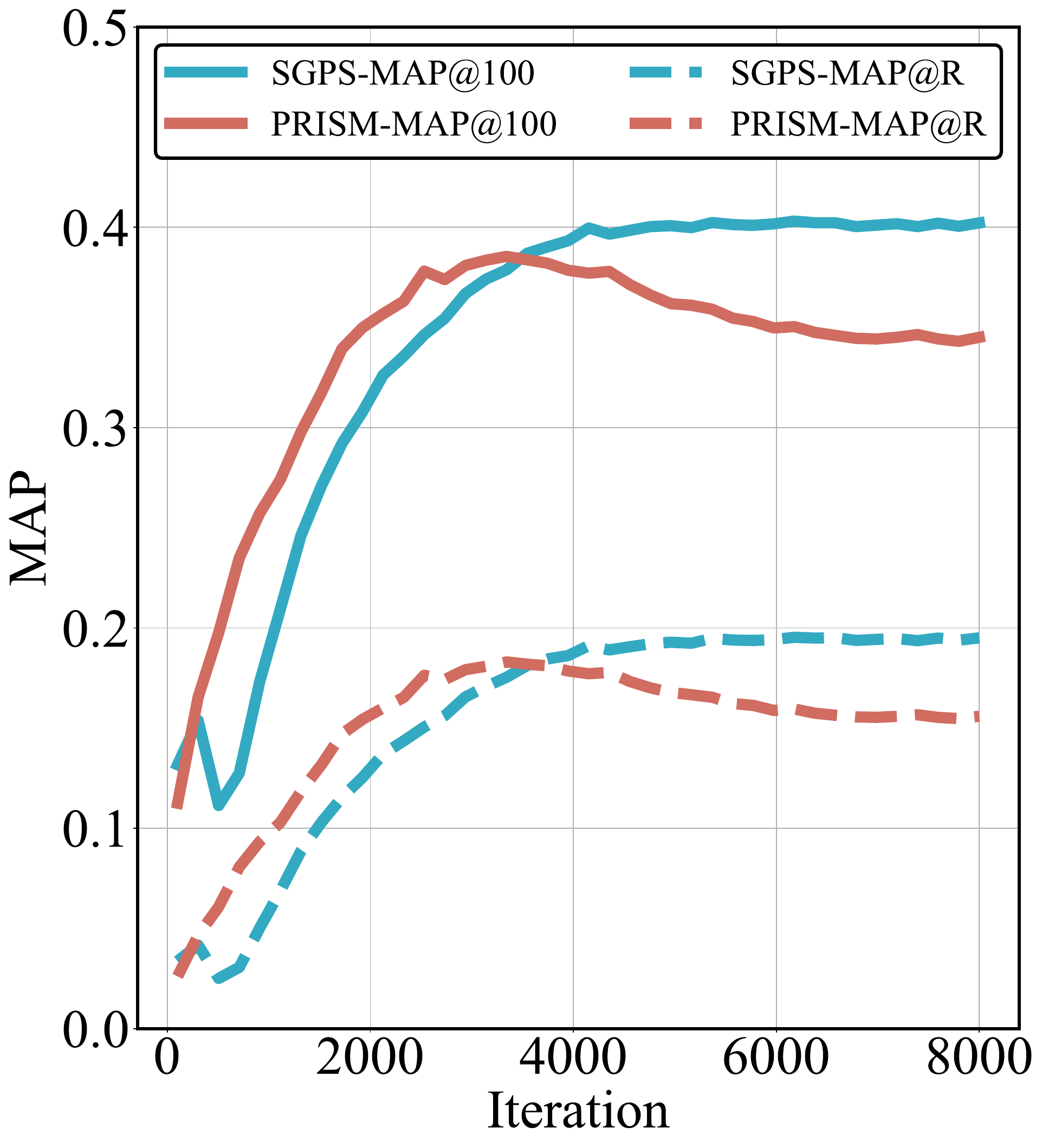}
    \label{fig:vis_map}
  }
  \caption{
    Change of Precision and MAP during the training.
    SGPS overcomes the overfitting problem in PRISM.
  }
  \label{fig:vis_change}
\end{figure}

\noindent\textbf{Overfitting Problems.}
We also investigate the overfitting problems of SGPS-MCL. As shown in Fig.\ref{fig:vis_precision} and Fig.\ref{fig:vis_map}, our method can achieve better performance than PRISM~\cite{PRISM2021} on all metrics, including Precision@1, R-Precision, MAP@R and MAR@100. Different from PRISM~\cite{PRISM2021}which suffers a performance degradation during the training, our method can keep a stable performance improvement, which indicates SGPS is less likely to overfit the noisy labels. In the training of SGPS, we can easily choose the best model when the training is finished, regardless of other cherry-picked strategies like early stopping.

\noindent\textbf{Runtime analysis.}
We provide the training speed comparisons for PRISM and SGPS ($K$=4) on CARS and SOP in Table\ref{tab:runtime}. It is indicated that the proposed model is about 1.1 times slower than PRISM on CARS (10\% noise) and SOP (10\% noise) and 1.2 times slower on CARS (50\% noise) and SOP (50\% noise). As we discussed before, the increased times are mainly caused by extra $K$ positive samples selected by SGM and the subsequent aggregating procedure in PPM. Due to the asynchronous update for $\bm{c}^{B},\bm{c}^{T}$, SGM will not add too much time costs during the training. Moreover, the costs during back-propagation and optimization are not affected.

\begin{table}[t]
  \centering
  \caption{GPU hours during the training. The speed is tested on an NVIDIA RTX 3090 GPU.}
  \label{tab:runtime}%
  \begin{tabular}{l|c|c}
    \toprule
    Dataset           & PRISM & SGPS  \\
    \midrule
    CARS (10\% noise) & 0.307 & 0.332 \\
    CARS (50\% noise) & 0.312 & 0.361 \\
    SOP (10\% noise)  & 3.090 & 3.368 \\
    SOP (50\% noise)  & 3.120 & 3.883 \\
    \bottomrule
  \end{tabular}%
\end{table}

\begin{figure}[!t]
  \begin{center}
    \includegraphics[width=1.0\linewidth]{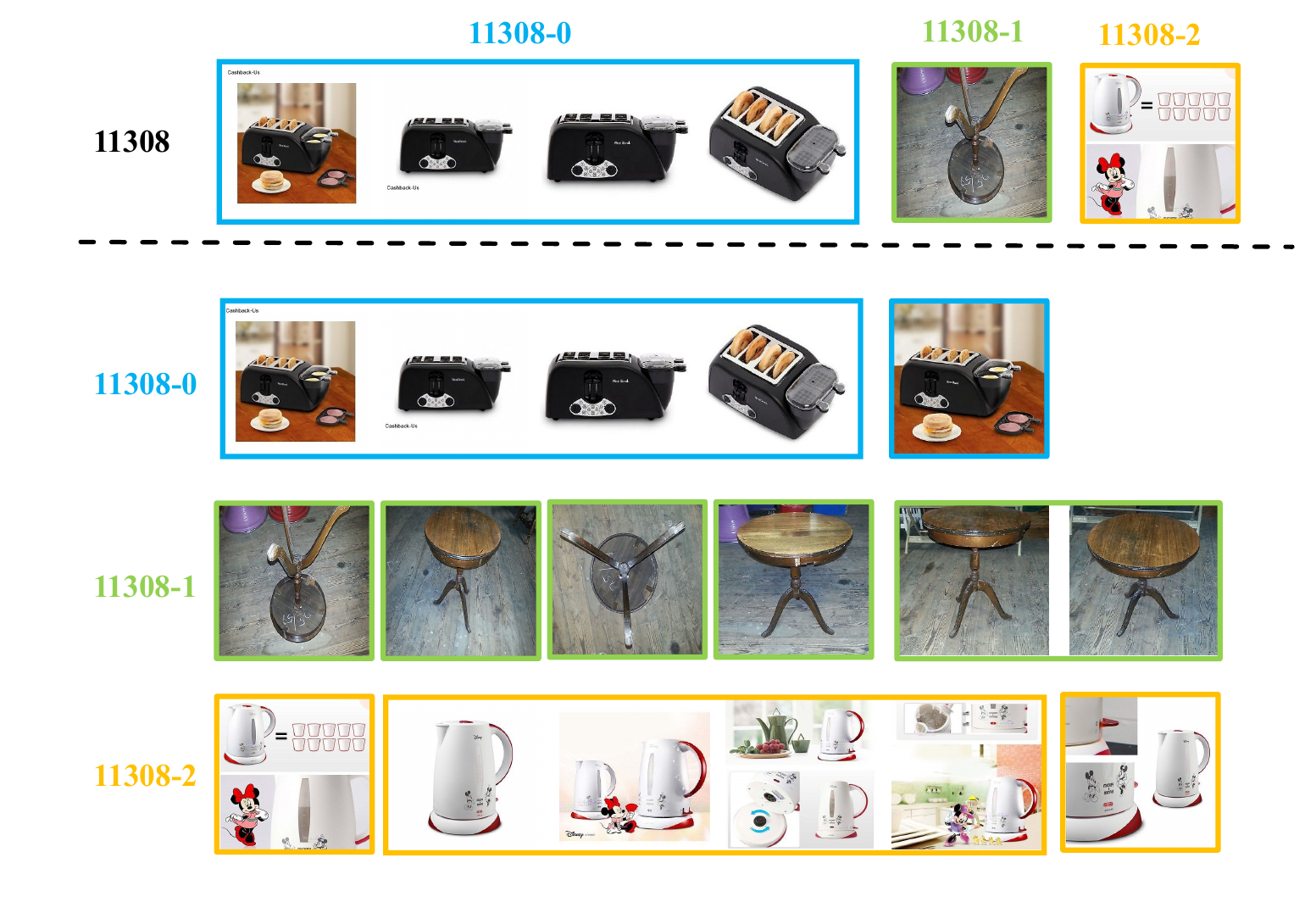}
    \caption{
      Visualization of SGM's results on SOP with 50\% symmetric noise. The first row shows some samples with the same noisy label (id 11308). The subsequent three rows show the samples from the three subgroups in $\bm{c}^B$.
      Samples within the same rectangle belong to the same subgroup and noisy label category.
      Different colors denote different ground truth categories.
    }
    \label{fig:vis_SOP}
  \end{center}
\end{figure}

\noindent\textbf{Discussion with multiproxy-based methods.}
\begin{figure}[!t]
  \begin{center}
    \includegraphics[width=0.7\linewidth]{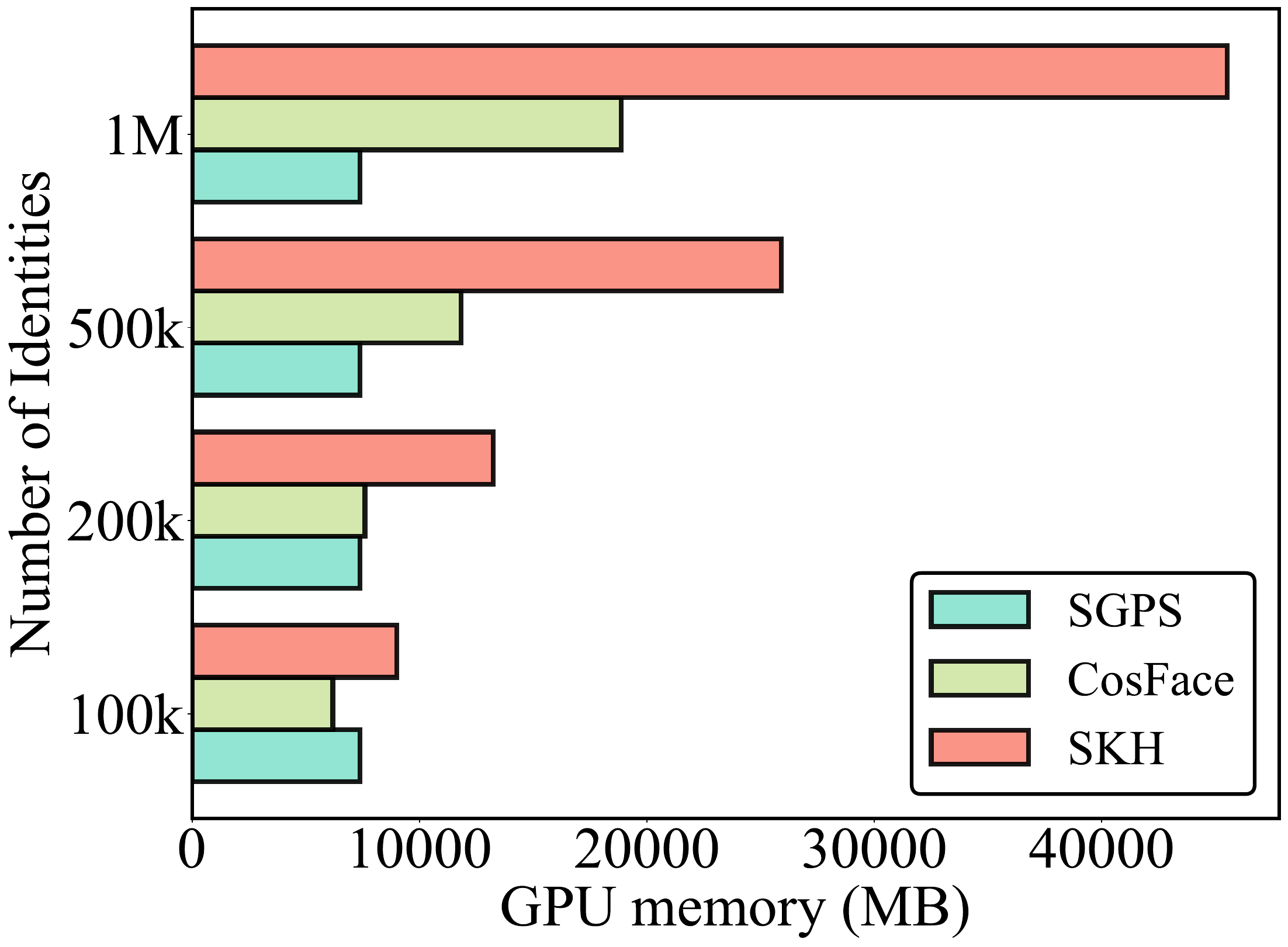}
    \caption{GPU memory cost of the multiproxy-based method and SGPS.}
    \label{fig:memory}
  \end{center}
\end{figure}
Multiproxy-based methods like SoftTriple~\cite{qian2019softtriple}, Subcenter~\cite{deng2020subcenter} and SKH~\cite{liu2021switchable} extend the softmax loss with multi centers for each class, which achieves SOTA performance on both DML benchmarks with clean labels and noisy labels. Our SGPS differs from the above multiproxy-based methods in two aspects. Firstly, SGPS is hardware-friendly and easy to incorporate with pair-wise methods like MCL~\cite{wang2020cross} and DCQ~\cite{li2021dynamic}. As shown in Fig.\ref{fig:memory}, multiproxy-based methods like SKH~\cite{liu2021switchable} cost several times more memory than CosFace~\cite{wang2018cosface}. In scenarios with over one million identities, the memory cost of SKH~\cite{liu2021switchable} will be unacceptable. Our SGPS will not add extra GPU memory cost during the training, which can be easily deployed on large-scale datasets. Secondly, SGPS introduces novel SGM to discover positive pairs instead of the predefined multiple centers. This allows SGPS to more effectively merge or filter out noisy samples and outliers than proxy-based methods. As illustrated in Fig. \ref{fig:vis_SOP}, when dealing with samples labeled as category 11308 in SOP with 50\% symmetric noise, the intra-class splitting process initially divides these samples into three subgroups: {11308-0, 11308-1, 11308-2}. Subsequently, after bottom-up subgroup generation, the resulting merged subgroups are displayed in the three lines below. It is noteworthy that samples within the same subgroup exhibit a high likelihood of belonging to the ground truth category (illustrated as the same color). Neverthelss, multiproxy-based methods are incapable of effectively modeling the noise situation like 11308-1, when positive samples are distributed across across numerous different categories. This example highlights the capability of SGPS in handling complex noise scenarios.

\section{Conclusion}
In this paper, we propose a novel SGPS framework for noise-robust DML. SGPS can be smoothly integrated with existing DML methods and notably reduces the impact of label noise, particularly in pair-wise DML approaches. This framework starts by efficiently distinguishing between clean and noisy samples using the PCS strategy. To enhance the utilization of noisy data, we introduce SGM to generate subgroup labels, which enable to SGPS discover positive pairs for noisy samples. Following that, PPM aggregates positive pairs into informative prototypes to involve the noisy samples in training process. Extensive experiments with both synthetic and real-world datasets demonstrate the effectiveness of SGPS.

\section*{Acknowledgments}
This work was supported in part by the National Key R\&D Program of China under Grant 2018AAA0102000, in part by National Natural Science Foundation of China: 62236008, U21B2038, U23B2051, 61931008, 62122075, 62406305, 62471013, 62476068 and 62272439, in part by Youth Innovation Promotion Association CAS, in part by the Strategic Priority Research Program of the Chinese Academy of Sciences, Grant No. XDB0680000, in part by the Innovation Funding of ICT, CAS under Grant No.E000000, in part by the China Postdoctoral Science Foundation (CPSF) under Grant No.2023M743441, and in part by the Postdoctoral Fellowship Program of CPSF under Grant No.GZB20230732.



%

\bibliographystyle{IEEEtran}
\bibliography{tip.bib}
\vfill

\begin{IEEEbiography}[{\includegraphics[width=1in,height=1.25in,clip,keepaspectratio]{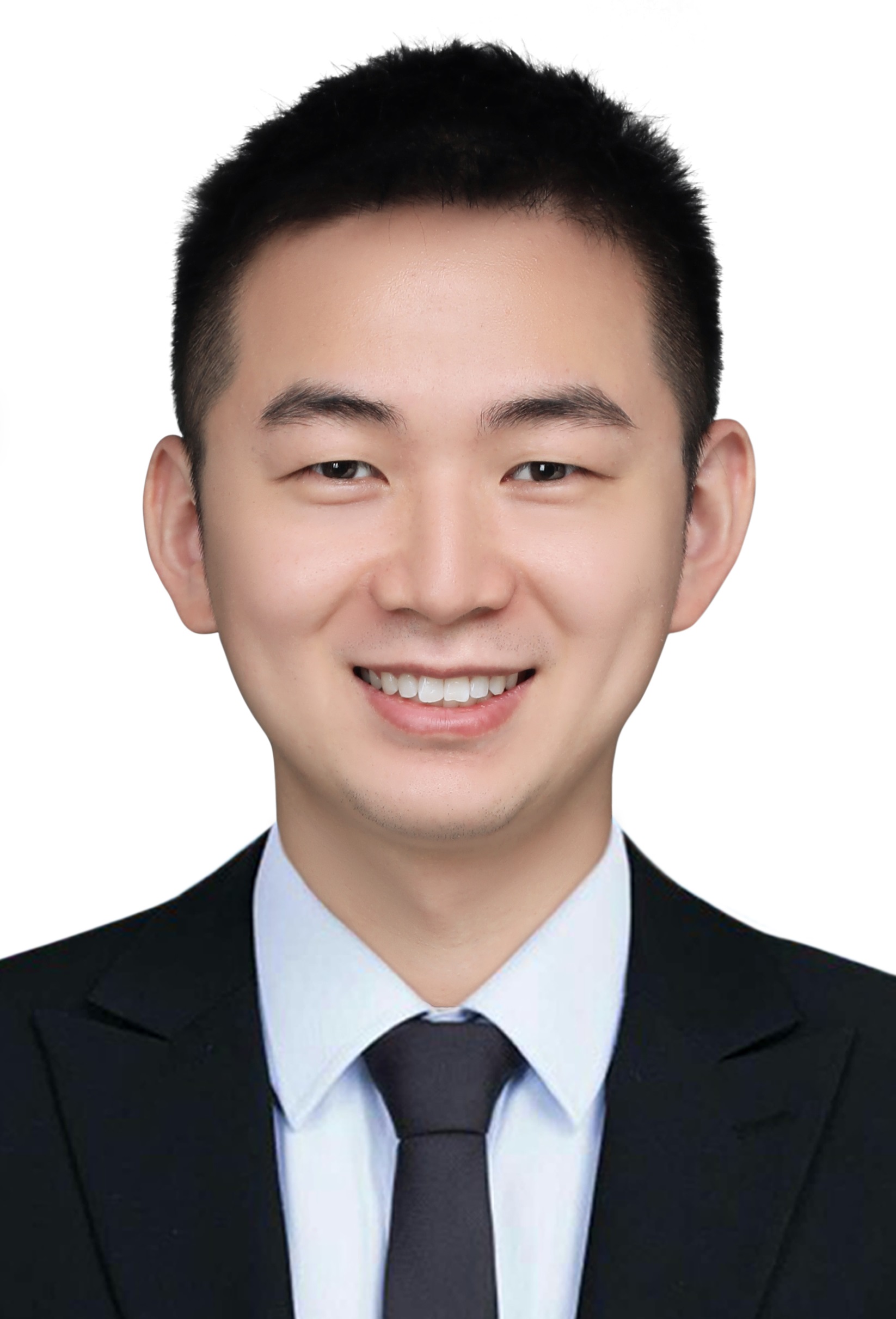}}]{Zhipeng Yu}
  received the B.E. degree in communication engineering and the M.E degree in electrical and communication engineering from
  the Beijing University of Posts and Telecommunications (BUPT), Beijing, China, in 2015 and 2018, respectively.
  He is currently pursuing the Ph.D. degree with  University of the Chinese Academy of Sciences.
  His research interests include machine learning and computer vision.
\end{IEEEbiography}

\begin{IEEEbiography}[{\includegraphics[width=1in,height=1.25in,clip,keepaspectratio]{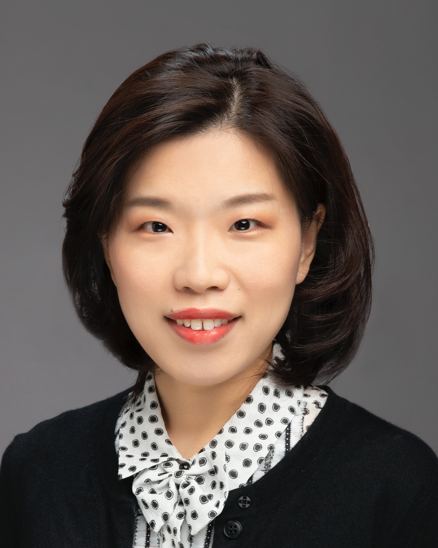}}]{Qianqian Xu}
  received the B.S. degree in computer science from China University of Mining and Technology in 2007 and the Ph.D. degree in computer science from University of Chinese Academy of Sciences in 2013. She is currently a Professor with the Institute of Computing Technology, Chinese Academy of Sciences, Beijing, China. Her research interests include statistical machine learning, with applications in multimedia and computer vision. She has authored or coauthored 90+ academic papers in prestigious international journals and conferences (including T-PAMI, IJCV, T-IP, NeurIPS, ICML, CVPR, AAAI, etc). Moreover, she serves as an associate editor of IEEE Transactions on Circuits and Systems for Video Technology, IEEE Transactions on Multimedia, and ACM Transactions on Multimedia Computing, Communications, and Applications.
\end{IEEEbiography}

\begin{IEEEbiography}[{\includegraphics[width=1in,height=1.25in,clip,keepaspectratio]{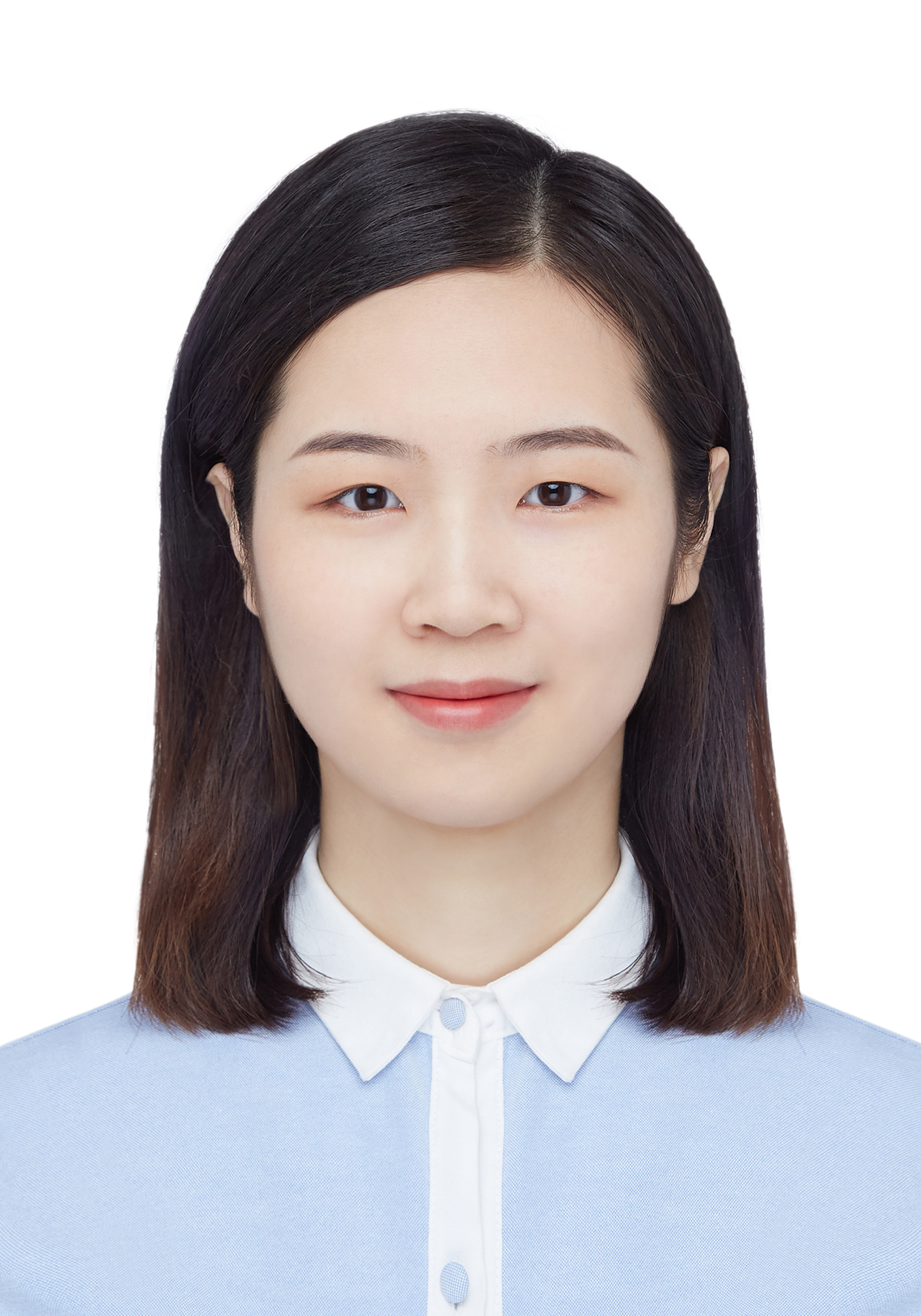}}]{Yangbangyan Jiang}
  received the B.S. degree in instrumentation and control from Beihang University in 2017
  and the Ph.D. degree in computer science from University of Chinese Academy of Sciences in 2023.
  She is currently a postdoctoral research fellow with University of Chinese Academy of Sciences.
  Her research interests include machine learning and computer vision.
  She has authored or coauthored several academic papers in international journals and conferences
  including T-PAMI, NeurIPS, CVPR, AAAI, ACM MM, {etc}.
  She served as a reviewer for several top-tier conferences such as ICML, NeurIPS, ICLR, CVPR, ICCV, AAAI.
\end{IEEEbiography}

\begin{IEEEbiography}[{\includegraphics[width=1in,height=1.25in,clip,keepaspectratio]{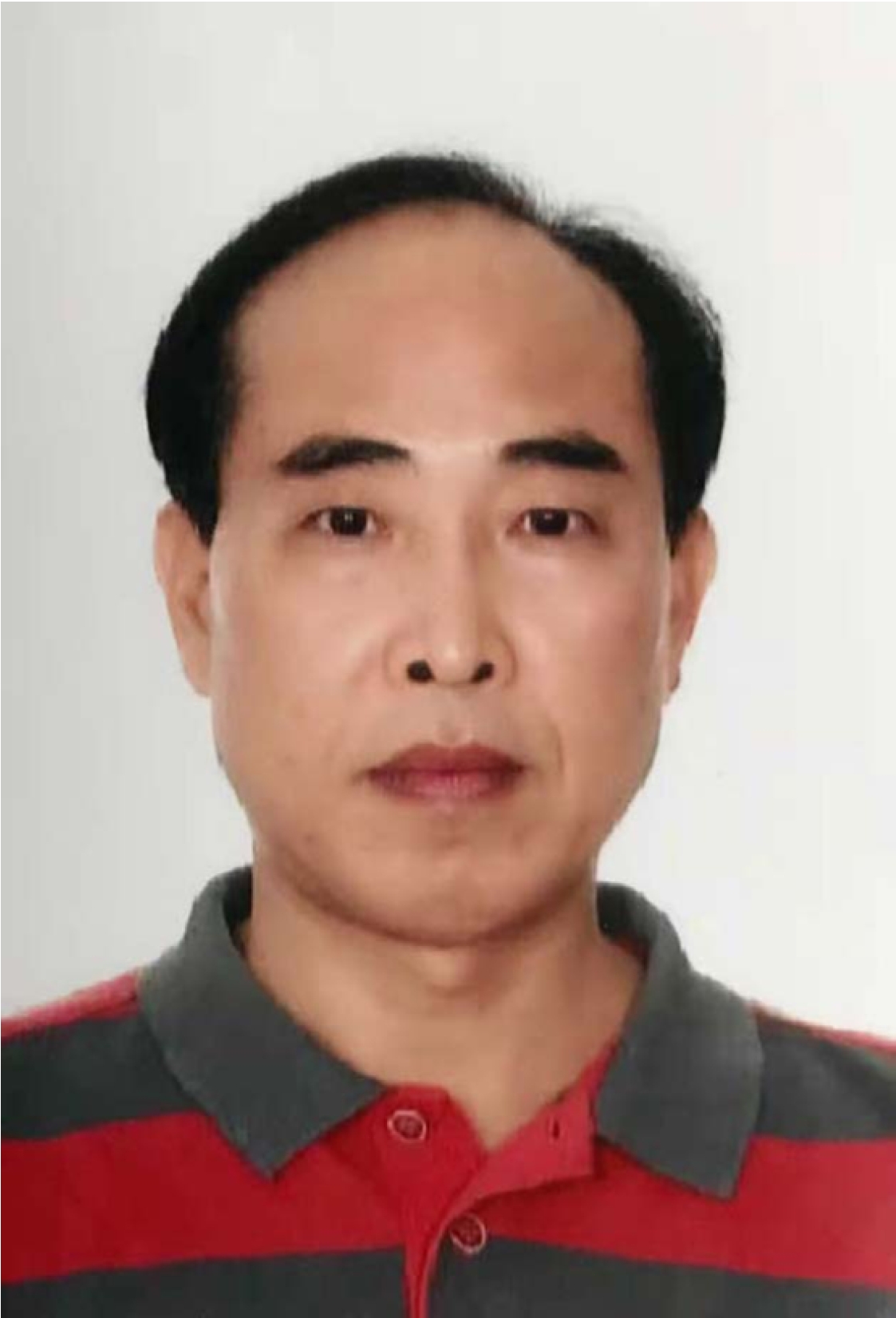}}]
  {Yingfei Sun} received the Ph.D. degree in applied
  mathematics from the Beijing Institute of
  Technology, in 1999. He is currently a Full Professor
  with the School of Electronic, Electrical
  and Communication Engineering, University of
  Chinese Academy of Sciences. His current research
  interests include machine learning and
  pattern recognition.
\end{IEEEbiography}

\begin{IEEEbiography}[{\includegraphics[width=1in,height=1.25in,clip,keepaspectratio]{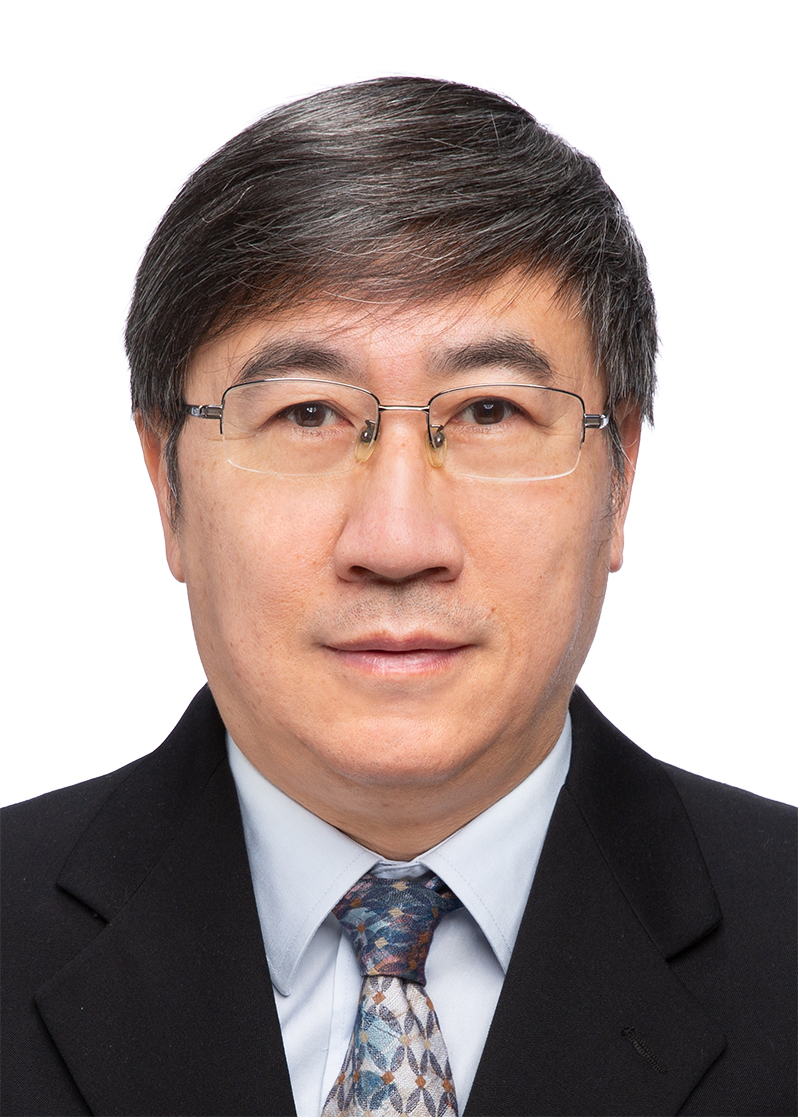}}]
  {Qingming Huang} is a chair professor in University of Chinese Academy of Sciences and
  an adjunct research professor in the Institute of Computing Technology,
  Chinese Academy of Sciences.
  He graduated with a Bachelor degree in Computer Science in 1988 and Ph.D. degree in Computer Engineering in 1994,
  both from Harbin Institute of Technology, China.
  His research areas include multimedia computing, image processing, computer vision and pattern recognition.
  He has authored or coauthored more than 400 academic papers in prestigious international journals and top-level international conferences.
  He was the associate editor of IEEE Trans. on CSVT and Acta Automatica Sinica, and the reviewer of various international journals including IEEE Trans. on PAMI, IEEE Trans. on Image Processing, IEEE Trans. on Multimedia, etc. He is a Fellow of IEEE and has served as general chair, program chair, area chair and TPC member for various conferences, including ACM Multimedia, CVPR, ICCV, ICME, ICMR, PCM, BigMM, PSIVT, etc.
\end{IEEEbiography}

\end{document}